\documentclass{ieeetj}
\usepackage{cite}
\usepackage{amsmath,amssymb,amsfonts}
\usepackage{algorithmic}
\usepackage{graphicx,color}
\usepackage{textcomp}
\usepackage{xcolor}
\usepackage{hyperref}
\hypersetup{hidelinks}
\usepackage{algorithm,algorithmic}
\def\BibTeX{{\rm B\kern-.05em{\sc i\kern-.025em b}\kern-.08em
    T\kern-.1667em\lower.7ex\hbox{E}\kern-.125emX}}
\AtBeginDocument{\definecolor{tmlcncolor}{cmyk}{0.93,0.59,0.15,0.02}\definecolor{NavyBlue}{RGB}{0,86,125}}

\usepackage{subcaption}
\usepackage{tabularx,booktabs}
\newcolumntype{C}{>{\centering\arraybackslash}X} 
\usepackage{xcolor,colortbl}
\definecolor{lavender}{rgb}{0.9, 0.9, 0.98}
\usepackage[capitalise]{cleveref}

\def\OJlogo{\vspace{-4pt}$<$Society logo(s) and publication title will appear here.$>$}
\def\seclogo{\vspace{10pt}$<$Society logo(s) and publication title will appear here.$>$}

\def\authorrefmark#1{\ensuremath{^{\textbf{#1}}}}

\begin{document}
\receiveddate{XX Month, XXXX}
\reviseddate{XX Month, XXXX}
\accepteddate{XX Month, XXXX}
\publisheddate{XX Month, XXXX}
\currentdate{XX Month, XXXX}
\doiinfo{XXXX.2022.1234567}

\markboth{}{Author {et al.}}

\title{Field evaluation and optimization of a lightweight autonomous lidar-based UAV system based on a rigorous experimental setup in boreal forest environments}

\author{Aleksi Karhunen\authorrefmark{1}, Teemu Hakala\authorrefmark{1}, Väinö Karjalainen\authorrefmark{1} and Eija Honkavaara\authorrefmark{1}}
\affil{Finnish Geospatial Research Institute in National Land Survey of Finland, Vuorimiehentie 5, 02150 Espoo, Finland}
\corresp{Corresponding author: Aleksi Karhunen (email: aleksi.karhunen@maanmittauslaitos.fi).}
\authornote{
This work has been submitted to the IEEE for possible publication. Copyright may be transferred without notice, after which this version may
no longer be accessible\\\\
This research was funded by the Research Council of Finland within project “DRONE4TREE - Autonomous drone solutions for single tree-based forest management” (decision no. 359404). This study has been performed with affiliation to the Research Council of Finland Flagship Forest–Human–Machine Interplay—Building Resilience, Re-defining Value Networks and Enabling Meaningful Experiences (UNITE) (decision no. 357908).}

\begin{abstract}
Interest in utilizing autonomous uncrewed aerial vehicles (UAVs) for under-canopy forest remote sensing has increased in recent years, resulting in the publication of numerous autonomous flight algorithms in the scientific literature. To support the selection and development of such algorithms, a reliable comparison of existing approaches based on published studies is essential. However, reliable comparisons are currently challenging due to widely varying experimental setups and incomplete reporting practices. This study proposes a standardized experimental setup for evaluating autonomous under-canopy UAV systems to fill this gap. The proposed setup emphasizes quantitative reporting of forest complexity, visual representation of test environments, execution of multiple repeated flights, and reporting of flight success rates alongside qualitative flight results. In addition, flights at multiple target speeds are encouraged, with reporting of realized flight speed, mission completion time, and point-to-point flight distance. 
The proposed setup is demonstrated using a lightweight lidar-based quadrotor employing state-of-the-art open-source algorithms, evaluated through extensive experiments in two natural boreal forest environments. Based on a systematic evaluation of the original system, several improvements were introduced. The same experimental protocol was then repeated with the optimized system, resulting in a total of 93 real-world flights. The optimized system achieved success rates of 12/15 and 15/15 at target flight speeds of 1 m/s and 2 m/s, respectively, in a medium-difficulty forest, and 12/15 and 5/15 in a difficult forest. Adoption of the proposed experimental setup would facilitate the literature-based comparison of autonomous under-canopy flight systems and support systematic performance improvement of future UAV-based forest robotics solutions.
\end{abstract}



\maketitle

\section{INTRODUCTION}

Interest in autonomous uncrewed aerial vehicles (UAVs) for under-canopy forest remote sensing has grown rapidly in recent years, driven by the increasing need for efficient, high-resolution characterization of forest structure and environment. Traditionally, under-canopy mobile remote sensing has relied on handheld or backpack-mounted systems as well as manually piloted UAV platforms \cite{9797818}. Although these approaches allow flexible data acquisition, they are constrained by limited scalability, high operator workload, and safety considerations. Autonomous quadrotors offer significant potential to improve data acquisition efficiency and enable access to environments that are difficult or unsafe for human operators. In response, several autonomous flight solutions have recently been developed, facilitating more efficient and scalable UAV-based under-canopy remote sensing.

Global Navigation Satellite System (GNSS) based positioning is unreliable beneath forest canopies due to multipath effects and signal blockages caused by vegetation \cite{schubert2010modeling}. Consequently, UAV operations in such environments must rely on onboard sensing to localize in unknown surroundings, perceive obstacles, plan safe trajectories, and avoid collisions. Autonomous flying solutions for GNSS-denied environments can be broadly categorized according to their sensor configurations and the algorithmic paradigms they employ. Most systems rely on lidar sensors \cite{liu2023integrated,ren2025safety,sharma2024autonomous,jacquet2025neural}, cameras \cite{zhou2022swarm,campos2021autonomous,xu2024quadcopter,tordesillas2021faster,huan2024uncertainty,zhang2025maplesscollisionfreeflightmpc,han2025dynamically} or a combination of both \cite{jacquet2025neural, Kurkutlu_Roohi_2025limpnet}. The traditional methods typically decompose the problem into separate modules for obstacle mapping, and path planning and control \cite{ren2025safety,liu2023integrated,zhou2022swarm,tordesillas2021faster,sharma2024autonomous,campos2021autonomous,xu2024quadcopter}. In the obstacle mapping stage, a three-dimensional representation, often a voxel or grid-based map of surrounding obstacles, is generated from sensor measurements. Subsequently, in path planning and control, a safe and feasible path is generated around the obstacles towards the goal, which the UAV then tracks. In recent years, deep-learning-based approaches have gained increasing attention  \cite{huan2024uncertainty,jacquet2025neural,Kurkutlu_Roohi_2025limpnet,han2025dynamically,delcol2025autonomous}. These methods often replace the separate obstacle mapping and path planning and control tasks with a deep neural network that directly maps the sensor observations to control commands and waypoints. In both traditional and learning-based systems, localization of the UAV is commonly performed using visual, lidar, or inertial odometry or simultaneous localization and mapping (SLAM) techniques. 

Reliable comparison of autonomous flight solutions is essential for selecting algorithms suitable for UAV-based under-canopy remote sensing applications. To demonstrate system performance in forest environments, simulation alone is insufficient; instead, autonomy must be evaluated under real-world forest conditions, with both reliability and effectiveness quantitatively assessed.  However, cross-study comparison remains challenging, as many published works lack sufficient rigor in experimental design or in the reporting of test environments and performance metrics. An analysis of performance evaluation metrics reported in relevant studies from the existing literature is presented in \autoref{sec:relatedWork}.

Therefore, this study proposes a standardized testing setup. The validity and usefulness of this setup for system development are demonstrated through a set of experiments conducted in accordance with the proposed protocol. The first set of experiments was carried out using a custom-built quadrotor equipped with a lightweight 3D lidar and openly available local planning algorithms. The original system was based on the autonomous flight algorithm IPC proposed by Liu et al. \cite{liu2023integrated} and the SLAM algorithm LTA-OM proposed by Zou et al. \cite{Zou20242455}. The initial experimental phase consisted of 33 flights conducted in two forest environments with varying levels of complexity. To enable direct comparison with the results reported by Liu et al. \cite{liu2023integrated}, the original system was used "as is", as published by Liu and Ren \cite{IPCsoftware} and HKU-Mars-Lab \cite{ROGmapSoftware, LTAOMsoftware}. The only modifications to the original code were those required to ensure compatibility between the different repositories. Based on the results and analysis of these initial 33 flights, the original algorithm was subsequently improved. The performance of the optimized system was evaluated through an additional 60 flights conducted in the same forest environments. In total, 93 real-world flights were conducted in this study. 

The main contributions of this study are summarized as follows:
\begin{itemize}
  \item This study identifies a lack of thorough experimental evaluation in the literature on autonomous flight algorithms. Insufficient experimentation and incomplete reporting make it difficult to reliably assess the performance of different methods. 
  \item This study proposes a standardized experimental setup for evaluating autonomous under-canopy flight systems, with the aim of enabling more rigorous and comparable assessment of different autonomous flight approaches. The validity and practical relevance of the proposed setup are demonstrated through comprehensive experiments conducted on a state-of-the-art autonomous flight system reported in the literature.
  \item An open-source, state-of-the-art autonomous under-canopy flight framework was deployed onboard a custom-built aerial robot and evaluated using the proposed testing setup. Based on systematic analysis of the experimental results, several improvements were introduced to the original algorithm to enable reliable autonomous flight in challenging forest environments. The resulting performance gains of the optimized system are validated through comparison with the original implementation and with relevant systems reported in the literature.
\end{itemize}

The rest of this study is structured as follows. \Cref{sec:relatedWork} reviews recent studies on real-world experiments conducted to evaluate the performance of autonomous under-canopy UAV flight systems. \Cref{sec:materialsAndMethods} describes the original and optimized algorithms, the hardware of the custom-built quadrotor, the test environments, and the performance assessment criteria used in this study. An overview of the experimental results is presented in \autoref{sec:results}. \Cref{sec:discussion} discusses the experimental results and presents and discusses the proposed standardized experimental setup in detail. The study is concluded in \autoref{sec:conclusion}.

\section{RELATED WORK}
\label{sec:relatedWork}

In the literature, the performance of autonomous flying systems for local point-to-point flying has been evaluated with varying degrees of experimental rigor. While simulation experiments are valuable for demonstrating feasibility and initial algorithmic performance, they are insufficient for assessing real-world operational robustness. Consequently, this section focuses on studies that evaluate autonomous flight systems through real-world forest experiments, as only such tests can reliably capture the practical challenges of under-canopy navigation. \Cref{tab:literatureTests} provides an overview of real-world forest experiments reported in recent studies, illustrating the level of experimental evaluation in the current literature.

\begin{table*}
    \caption{Overview of real-world forest experiments reported in prior studies. "x" indicates that the corresponding element was performed and reported, while "-" indicates that the element was omitted. "$\approx$" denotes values that were not explicitly reported but were approximated based on the information available in the study. * Indicates that values were not reported in the original study but approximated by the authors of this paper based on figures and supplementary materials.} 
    \label{tab:literatureTests}
    \begin{tabularx}{\textwidth}{@{} p{1.9cm} *{11}{C} @{}}
    \toprule
    Study & Real forest flight & Density & Number of low-hanging branches & Flight distance & Multiple flights & Success rate & Collisions not leading to a failure & Target flight speed & Actual flight speed & Flight completion time\\
    \midrule
    Campos-Macías et al. \cite{campos2021autonomous} (2021) & x & - & - & 11.58 m & - & - & - & - & 0.2 m/s & 26.4 s \\
    \rowcolor{lavender}
    Zhou et al. \cite{zhou2022swarm} (2022) & x & High* & Low* & 65 m & - & - & - & - & - & - \\ 
    Tordesillas et al. \cite{tordesillas2021faster} (2022) & - & - & - & - & - & - & -& -& - & - \\
    \rowcolor{lavender}
    Karjalainen et al. \cite{karjalainen2023drone} (2023) & x & 1650 -- 2380 trees/ha & High & 35 -- 80 m & 19 & 47\% & - & 1.0 m/s & - & - \\
    Liu et al. \cite{liu2023integrated} (2024) & x & Low* & Low* & 58.64 m & $>$10 & - & - & 1.0 -- 6.0 m/s & 4.51 m/s (for the 6 m/s flight) & - \\
    \rowcolor{lavender}
    Sharma \& Liang \cite{sharma2024autonomous} (2024)  & - & - & - & - & - & - & -& -& - & - \\
    Xu \& Shimada \cite{xu2024quadcopter} (2024) & - & - & - & - & - & - & -& -& - & - \\
    \rowcolor{lavender}
    Nguyen et al. \cite{huan2024uncertainty} (2024) & x & 2000 trees/ha & Medium* & $\approx$60 m & - & - & - & 1.5 m/s & - & - \\    
    Ren et al. \cite{ren2025safety} (2025) & Partly & Open area and Low* & Low* & $\approx >$550 m & 8 & 100\% & - & 5 -- 20 m/s & 5 -- 12.5 m/s  & - \\
    \rowcolor{lavender}
    Jacquet et al. \cite{jacquet2025neural} (2025) & x & Medium* & - & 15 m & - & - & - & 1.5 m/s & $\approx$ 1.25 m/s & - \\
    Zhang et al. \cite{zhang2025maplesscollisionfreeflightmpc} (2025) & - & - & - & - & - & - & -& -& - & - \\
    \rowcolor{lavender}
    Han et al. \cite{han2025dynamically} (2025) & x & 500 trees/ha & Low* & $>$50 m & 10 & - & - & 5 m/s & 3.7 m/s & - \\
    Kurkutlu \& Roohi \cite{Kurkutlu_Roohi_2025limpnet} (2025) & - & - & - & - & - & - & -& -& - & - \\
    \rowcolor{lavender}
    Del Col et al. \cite{delcol2025autonomous} (2025) & x & -/high* & Moderate/ high & $\approx$15/15 -- 80 m & 5/4 & 100\%/50\% & Some/none & -/0.8 -- 1.0 m/s & - & - \\
    Karjalainen et al. \cite{Karjalainen03122025} (2025) & x & 650/2000 trees/ha & High/ high & 34--36 / 42 m & 7/9 & 100\%/89\% & - & 1.0 m/s & - & - \\ 
    \bottomrule
    \end{tabularx}
\end{table*}

Campos-Macías et al. \cite{campos2021autonomous} reported a single 12 m autonomous flight conducted in a forest environment, for which no information on forest complexity was provided. Based on the available imagery and the obstacle map presented in the study, the complexity of the forest environment could not be reliably estimated. In addition to the outdoor experiment, three further real-world experiments were conducted in indoor settings. In the first experiment, the system navigated through a maze constructed from boxes. In the second experiment, a 29 m flight was conducted inside a warehouse over a duration of 160 s. In the final experiment, the system performed a 5 m flight while avoiding walking obstacles. 

Zhou et al. \cite{zhou2022swarm} conducted all experiments using multiple quadrotors as a swarm. In the forest flight experiment, the swarm completed a single 65-meter flight. The density of the test forest was not explicitly reported, but based on the provided imagery, the environment can be qualitatively assessed as having high tree density with relatively few low-hanging branches. In addition to the forest experiment, three further real-world experiments were conducted. The first involved a single flight through an artificial forest, during which the swarm maintained a predefined formation. In the second experiment, all ten quadrotors were assigned random goal points distributed along a circle with a radius of 3 m to evaluate inter-robot collision avoidance. In the third experiment, the swarm was tasked to track a human through a wooded area under high computational load. 

Liu et al. \cite{liu2023integrated} reported more than ten successful real-world flights with target flight speeds ranging from 1.0 to 6.0 m/s. However, the overall success rate was not quantified, as the study does not mention whether any failed flights occurred during testing. The complexity of the forest was not reported. Based on the provided imagery of the test site, the forest can be assessed as having low density with a low amount of low-hanging branches. In addition to the real-world forest experiments, Liu et al. conducted a series of tests evaluating the system’s ability to avoid suddenly appearing obstacles. In addition, indoor experiments were performed in which the quadrotor flew a star-shaped path in the presence of dynamic obstacles. In the final set of experiments, the quadrotor was subjected to external disturbances, including wind- and contact-based force disturbances applied using a stick. 

Nguyen et al. \cite{huan2024uncertainty} reported a single 60 m autonomous flight conducted in a forest environment with a reported density of 2000 trees/ha and a qualitatively estimated moderate amount of low-hanging branches. The target flight speed was 1.5 m/s. The other real-world experiments were conducted in indoor environments. In the first indoor experiment, the robot completed a 35 m flight through a cluttered corridor. In the second experiment, the system navigated inside a silo tank along a predefined path while deviating from the given path when an object of interest was detected. The final experiment followed a similar setup but was conducted inside a hall of a campus building instead of a silo tank.

Ren et al. \cite{ren2025safety} conducted three real-world experiments. In the forest flight experiment, the quadrotor was flown eight times through a test area composed primarily of open spaces and a low-complexity forest environment with large gaps between trees. The flights were conducted under varying illumination conditions ranging from daylight to complete darkness. The density of trees in the test environment was not reported. The target flight velocities ranged from 5 to 20 m/s, and the success rate was reported, with all flights completed successfully. The actual flight speed ranged between 5 m/s with a target flight speed of 5 m/s and 12.5 m/s with a target flight speed of 20 m/s. In addition to forest experiments, the system was evaluated in a person-tracking task within an environment consisting of tree trunks without branches and leaning logs. Finally, the performance of the system was assessed in a thin-wire avoidance task and compared against a commercial UAV. 

Jacquet et al. \cite{jacquet2025neural} reported a single 15-meter autonomous flight conducted in a forest environment, for which the tree density was not reported. Based on the provided point cloud of the test site, the forest density can be qualitatively assessed as low, although the presence of low-hanging branches and understory foliage could not be reliably estimated. In addition to forest flight, two further experiments were conducted. The first consisted of a 35 m flight in which the reference path was provided by a human operator who guided the quadrotor towards the surrounding obstacles. The final experiment involved a 35-meter flight performed under drifting odometry.

Han et al. \cite{han2025dynamically} reported a single real-world experiment consisting of ten autonomous flights conducted in a forest environment with a density of 500 trees/ha. The success rate of these flights was not reported. The presence of low-hanging branches and other understory foliage was qualitatively assessed as low based on the provided imagery. All flights exceeded 50 m in length and were flown with a target speed of 5 m/s.

Del Col et al. \cite{delcol2025autonomous} conducted multiple flights in two forest environments. The densities of the forests were not reported numerically; instead, a qualitative scale of three-levels was used. The first environment was a small 15$\times$15 m² forest patch, described as having a "medium" density. Based on the provided imagery, the density could not be reliably verified, although a moderate number of low-hanging foliage was observed. The second environment was a larger forested area, reported to have a "high" density, which appeared to be consistent with the images. The flight distances were 15 m in the first test environment, and 15, 30, 60, and 80 m in the second. In the first environment, five flights were completed with a success rate of 100 \%, whereas in the second environment, four flights were conducted with a success rate of 50 \%. There were a few collisions with obstacles in the first set of experiments that did not lead to failures, and no collisions were reported in the second. The target flight velocities for the second environment flights were 0.8 or 1.0 m/s. 

Sharma and Liang \cite{sharma2024autonomous}, Xu and Shamada \cite{xu2024quadcopter}, Tordesillas et al. \cite{tordesillas2021faster}, and Kurkutlu and Roohi \cite{Kurkutlu_Roohi_2025limpnet} did not report any real-world forest flight experiments. Sharma and Liang \cite{sharma2024autonomous} and Kurkutlu and Roohi \cite{Kurkutlu_Roohi_2025limpnet} conducted evaluations in simulation environments. Xu and Shamada \cite{xu2024quadcopter} reported real-world experiments only in indoor settings, consisting of two different box-based environments. Similarly, Tordesillas et al.\cite{tordesillas2021faster} conducted six indoor flights through a box-filled obstacle course, in which either the goal locations or the box configuration were modified between flights. 

Karjalainen et al. \cite{karjalainen2023drone,Karjalainen03122025} conducted multiple studies using an autonomous flight system based on the algorithm proposed by Zhou et al. \cite{zhou2022swarm}. In the main experiment \cite{karjalainen2023drone}, a total of 19 flights were flown in a forest environment comprising two distinct patches. The first patch, 35 m long, had a reported density of 2380 trees/ha, and the second patch had a reported density of 1650 trees/ha. Both forests were described to have a high number of low-hanging branches, although the second patch had slightly fewer. The target flight velocities were 1.0 m/s, and the flight distances ranged from 35 to 80 m. The overall success rate for these flights was 47 \%. Two additional real-world experiments were conducted in a sparse mixed forest and in a park woodland with understory vegetation; tree densities were not reported for these environments. In the sparse mixed forest, eight of nine flights were successful with target velocities between 0.5 and 1.0 m/s. In the park woodland, three of the nine flights succeeded with the target velocities between 1.0 and 1.5 m/s.

Although the primary focus of Karjalainen et al. \cite{Karjalainen03122025} was the use of an autonomous quadrotor for DBH (diameter at breast height) measurements, the flight performance of the improved system from \cite{karjalainen2023drone} was evaluated. A total of 16 flights were conducted in two forest environments with reported densities of 650 trees/ha and 2000 trees/ha. While the amount on low-hanging branches and understory foliage was not directly reported, the forest complexity criterion proposed by Liang et al. \cite{liang2019forest} was used to categorize the forests. Based on this criterion, both forests were classified as having a high number of low-hanging branches and other foliage. Despite the low density of the medium forest, the high number of low-hanging branches and other foliage made the forest more complex than the density suggests. The target flight speed was 1.0 m/s in all test flights. The flight distances ranged from 34 to 36 m in the first forest and were 42 m in the second forest. All seven flights in the first forest were successful, while eight out of nine flights succeeded in the second forest. 

Overall, these studies highlight the need for more standardized experimental setups and reporting when proposing new autonomous flight solutions for under-canopy forest environments. Significant variation in testing setups and reporting across the literature makes performance comparisons between systems difficult. In many studies, key metrics such as success rate and forest complexity are not reported, making it nearly impossible to accurately assess the reliability of the proposed systems. To address these limitations, this paper proposes a standardized experimental setup and emphasizes consistent reporting of key performance metrics to enable transparent and comparable evaluation. 

\section{MATERIALS AND METHODS}
\label{sec:materialsAndMethods}

\subsection{ORIGINAL AUTONOMOUS FLIGHT ALGORITHM}

Two open-sourced algorithms, IPC by Liu et al. \cite{liu2023integrated}, responsible for path planning and control, and LTA-OM by Zou et al. \cite{Zou20242455}, a SLAM algorithm responsible for the localization of the quadrotor, served as the foundation for the original system. The obstacle mapping module of IPC utilizes a simplified version of ROG-Map \cite{ren2023rogmap}, where a voxel is marked occupied if it is hit by a lidar scan. Liu et al. introduced temporal forgetting to ROG-Map, where a voxel is marked unoccupied if no lidar scan has hit the voxel during the forgetting threshold. A* is used to find the shortest path towards the goal, and a reference path is created by pruning away redundant nodes to generate the shortest piece-wise reference path. 

The model predictive control (MPC) and the differential flatness property of the quadrotors \cite{fliess1995flatness} are used to generate a set of control inputs for the flight control unit (FCU). In the source code of IPC provided by Liu et al. \cite{IPCsoftware}, the MPC problem differs slightly from the one given in the original article \cite{liu2023integrated} and is formulated as
\begin{subequations}
\label{eq:mpcProb}
\begin{alignat} {1}
\label{eq:mpcProba}
    &\min_{\mathbf{u}_k} \sum^N_{n=1}||\mathbf{u}_{n-1}||^2_{\mathbf{R}_u} + \sum^{N-1}_{n=1}\big(||(\mathbf{p}_{\text{ref}, n} - \mathbf{p}_n)||^2_{\mathbf{R}_p} \notag \\ 
    &+ ||(\mathbf{v}_{\text{ref}, n} - \mathbf{v}_n)||^2_{\mathbf{R}_v} + ||\mathbf{a}_n||^2_{\mathbf{R}_a}\big) + ||(\mathbf{p}_{\text{ref}, N} - \mathbf{p}_N)||^2_{\mathbf{R}_{p, N}} \notag\\ 
    &+ ||\mathbf{v}_N||^2_{\mathbf{R}_{v, N}} + ||\mathbf{a}_N||^2_{\mathbf{R}_{a, N}} + \sum^{N-2}_{n=0} ||\mathbf{u}_{n+1} - \mathbf{u}_n||^2_{\mathbf{R}_c} 
\end{alignat}
\begin{alignat} {2}
        \textbf{s.t.} \quad 
        \label{eq:MPCconstraintb}
        \mathbf{x}_n = \mathbf{f}_d(\mathbf{x}_{n-1}, \mathbf{u}_{n-1}), \quad n = 1,2,\dots,N
    \end{alignat}
    \begin{alignat} {3}
        \label{eq:MPCconstraintc}
        \mathbf{x}_0 = [\mathbf{p}_{\text{odom}}, \mathbf{v}_{\text{odom}}, \mathbf{a}_{\text{odom}}]^T
    \end{alignat}
    \begin{alignat} {4}
        \label{eq:MPCconstraintd}
        v_{i\text{, min}} \leq v_{i, n} \leq v_{i\text{, max}}, \quad i=x,y,z
    \end{alignat}
    \begin{alignat} {5}
        \label{eq:MPCconstrainte}
        a_{i\text{, min}} \leq a_{i, n} \leq a_{i\text{, max}}, \quad i=x,y,z
    \end{alignat}
    \begin{alignat} {6}
        \label{eq:MPCconstraintf}
        j_{i\text{, min}} \leq j_{i, n} \leq j_{i\text{, max}}, \quad i=x,y,z
    \end{alignat}
    \begin{alignat} {7}
        \label{eq:MPCconstraintg}
        \mathbf{C}_n \cdot \mathbf{p}_n - \mathbf{d}_n \leq 0.
\end{alignat}
\end{subequations}
In \eqref{eq:mpcProba}, $\mathbf{R}_u$ and $\mathbf{R}_c$ are the weight parameters for current control and control variation. $||\mathbf{u}_{n-1}||^2_{\mathbf{R}_u}$ is the control effort given as the jerk and $||\mathbf{u}_{n+1} - \mathbf{u}_n||^2_{\mathbf{R}_c}$ is the control variation. $\mathbf{R}_p$, $\mathbf{R}_v$, and $\mathbf{R}_a$ are the intermediate weight parameters for position, velocity, and acceleration. $||(\mathbf{p}_{\text{ref}, n} - \mathbf{p}_n)||^2_{\mathbf{R}_p}$, $||(\mathbf{v}_{\text{ref}, n} - \mathbf{v}_n)||^2_{\mathbf{R}_v}$, and $||\mathbf{a}_n||^2_{\mathbf{R}_a}$ are the reference path, velocity, and acceleration following an error at the reference positions $[\mathbf{p}_{\text{ref},1}, \ldots,  \mathbf{p}_{\text{ref},N-1}]$. $\mathbf{R}_{p, N}$, $\mathbf{R}_{v, N}$, and $\mathbf{R}_{a, N}$ are the weight parameters for position, velocity, and acceleration at the horizon length $N$. $||(\mathbf{p}_{\text{ref}, N} - \mathbf{p}_N)||^2_{\mathbf{R}_{p, N}}$, $||\mathbf{v}_N||^2_{\mathbf{R}_{v, N}}$, and $||\mathbf{a}_N||^2_{\mathbf{R}_{a, N}}$ are the reference path, velocity, and acceleration error at the reference position of the horizon length $N$. The MPC problem is practically the same as that presented in \cite[eq. (2a)]{liu2023integrated}, but with greater flexibility in parameter selection. Instead of having the same weight parameter for every position, velocity, and acceleration index, a different weight parameter can be used for the intermediate reference indexes and the last index. 

The MPC problem \eqref{eq:mpcProba} is constrained by equations \eqref{eq:MPCconstraintb}--\eqref{eq:MPCconstraintg}. Equation \eqref{eq:MPCconstraintb} constraints the state $\mathbf{x}_n$ to the previous state $\mathbf{x}_{n-1}$ and the previous control input $\mathbf{u}_{n-1}$ using a third-order integrator. Equation \eqref{eq:MPCconstraintc} states that the initial state $\mathbf{x}_0$ is estimated by odometry $[\mathbf{p}_{\text{odom}}, \mathbf{v}_{\text{odom}}, \mathbf{a}_{\text{odom}}]^T$. Equations \eqref{eq:MPCconstraintd}--\eqref{eq:MPCconstraintf} constrain the velocity $v_{i, n}$, acceleration $a_{i, n}$, and jerk $j_{i, n}$ between the maximum and minimum values for velocity, acceleration, and jerk. Equation \eqref{eq:MPCconstraintg} constraints the position to lie inside the two safe flight corridors (SFC.) \cite{liu2017planning}. ($\mathbf{C}_n$, $\mathbf{d}_n$) represents the set of hyperplanes that form one of the polyhedra of the $n$\textsuperscript{th} SFC.

The MPC problem is solved with OSQP-Eigen \cite{OSQP-Eigen}. If the MPC problem could not be solved, the corresponding control command from the last successful solve is used instead. The control commands given by solving the MPC problem, given as a set of jerk commands, are transformed to a set of attitude commands utilizing the differential flatness property \cite{fliess1995flatness} of quadrotors. 

LTA-OM utilizes FAST-LIO2 \cite{xu2022fast} as the odometry module, and STD-LCD \cite{yuan2023std} is used to detect loop closures. LTA-OM employs a false positive rejection scheme where every loop closure candidate has to pass a consistency check before being inserted properly into the pose graph. FAST-LIO2 of LTA-OM was modified so that odometry is updated at the update rate of the IMU (Inertial Measurement Unit) instead of the update rate of the lidar rate. Between the odometry estimations from the lidar scans, the odometry estimation is forward propagated from the inertial measurements with a second-order integrator. The whole system operates on the Robot Operating System (ROS) \cite{quigley2009ros}.

\subsection{OPTIMIZED AUTONOMOUS FLIGHT ALGORITHM}

Based on the experiments conducted with the original system, several modifications were implemented to the autonomous flying algorithm.

During the flight experiments with the original system, the quadrotor exhibited local oscillatory behavior, particularly in denser forest areas. Since A* always seeks the shortest path toward the goal, re-planning could result in a new reference path in a completely different direction than the previous one. This often caused abrupt changes in flight direction, leading to oscillatory motions such as repeated vertical or lateral movements, with the quadrotor remaining nearly stationary for extended periods. A more detailed description of the behavior is presented in \autoref{subsec:descriptionFailures}. To mitigate this issue, the path planning module was modified so that A* tries to generate a path that follows the last reference path during the re-planning process. This is achieved by modifying the heuristic function of A* with an additional penalty term that penalizes the grid nodes based on their distance from the previously generated reference path. This modification biases the planner toward solutions that are more consistent with the prior path, resulting in smoother and more stable flight behavior:
\begin{equation}
    \label{eq:AstarFollowHeuristic}
    h(n) = 
    \begin{cases}
        d(n)_{\text{goal}} + w d(n)_{\text{last\_path}}& d(n)_{\text{start}} \leq d_{\text{follow}} \\
        d(n)_{\text{goal}}& d(n)_{\text{start}} > d_{\text{follow}}
    \end{cases},
\end{equation}
where $d(n)_{\text{goal}}$ is the distance from the current grid node $n$ to the goal, $d(n)_{\text{last\_path}}$ is the shortest distance from $n$ to the last reference path, $d(n)_{\text{start}}$ is the distance from $n$ to the starting grid node of the search, $d_{\text{follow}}$ is the distance threshold from the start inside of which the old path should be followed, and $w$ is the weight parameter which influences how strongly the old path should be followed. The modified heuristic can significantly slow down the A* search when a feasible path toward the goal is difficult to find. To mitigate this, after the first node is found outside the old path following distance $d_{\text{follow}}$, neighbors that are located closer than 80\% of $d_{\text{follow}}$ from the start are not added to the open set. This might lead to a situation where a path is not found, even when one exists; hence, a new search is triggered without the old path following if no path is found. In some cases, the A* search might still require excessive computation time. Therefore, an emergency stop procedure is activated if the path search takes longer than 0.1 s. After triggering the emergency stop, the same A* search is resumed incrementally in 0.1 s time slices until the search terminates. 

The second modification addresses error handling when "Not a number" values (NaNs) are generated during the SFC generation process. If NaN values are not properly handled, the MPC may produce control inputs containing NaNs, which in turn cause the PX4 FCU used in the system to shut down the motors. In the optimized system, the occurrence of NaNs triggers an emergency stop procedure. During this procedure, all MPC target positions are set to the current vehicle position, target velocities and accelerations are set to zero, and the SFC is replaced with a default bounding box as described by Liu et al. \cite{liu2017planning}.

Lastly, the direction of gravity is measured before takeoff by taking 200 IMU measurements and taking the average direction of acceleration as the direction of gravity. This enables the quadrotor to plan paths perpendicular to the direction of gravity, even when the takeoff location is on sloped terrain. Without this initial gravity alignment, height estimation becomes inaccurate when the flight is initiated from a slope.

\subsection{HARDWARE}

The experiments were conducted with a custom-built quadrotor. The onboard computer was a Minisforum EM780 (Minisforum, Hong Kong, China) with an AMD Ryzen 7 7840U (AMD, Santa Clara, CA, USA) CPU. Pixhawk 6C mini (Holybro, Hong Kong, China) was used as the PX4-compliant FCU, which was responsible for low-level control of the quadrotor given the attitude commands generated by the MPC module. As the 3D lidar of the system, Livox Mid-360 (Livox Technology, Shenzhen, China) was selected. It is a lightweight and small lidar, weighing 265 g, with an exterior measure of 65$\times$65$\times$60 mm, which enables the whole system to be similar in weight and size to robotic quadrotors based on stereo cameras. Livox Mid-360 outputs 200,000 points a second in a non-repeating manner, and the field of view of the point cloud is 360\textdegree \ horizontally, 7\textdegree \ downwards, and 52\textdegree  \ upwards. The built-in IMU of the Livox Mid-360, ICM40609, was utilized as the IMU of the system. The publishing rates were set to 10~Hz for the point cloud and 200~Hz for the IMU. 

\begin{figure}[tb]
    \includegraphics[height=6.0cm]{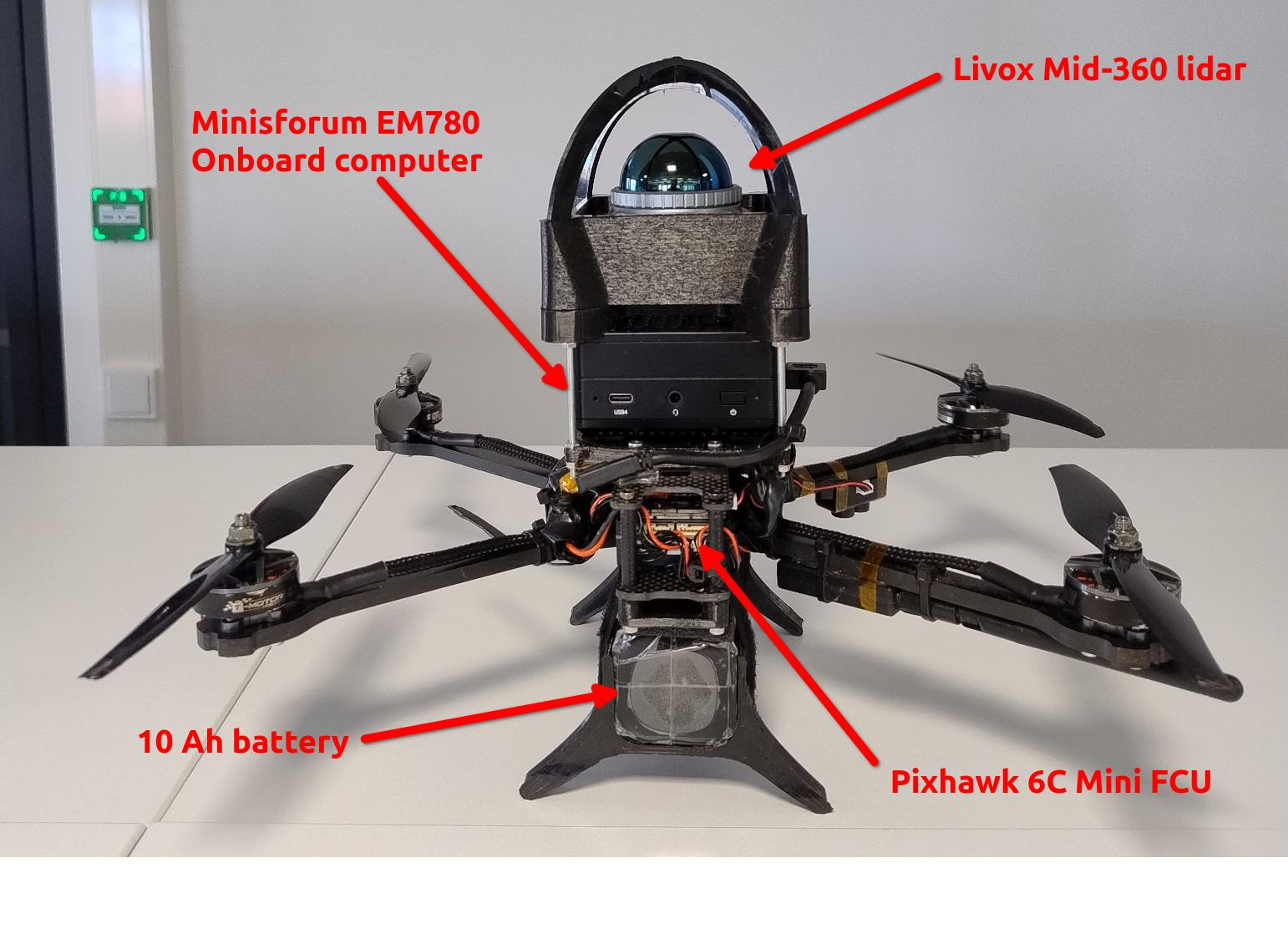}
    \caption{The custom-built quadrotor used in the experiments.}
    \label{fig:quadrotor}
\end{figure}

The hardware was mounted on a quadrotor frame with a motor-to-motor distance of 350 mm and equipped with two-bladed propellers with a length of 17.8 cm (8 inches). The platform weighed 1245 g without the battery and 1875 g with the 10 Ah battery installed. The quadrotor platform is shown in \autoref{fig:quadrotor}.

\subsection{TEST ENVIRONMENTS AND EXPERIMENTAL SETUP}

\begin{figure}[tb]
\begin{subfigure}{0.48\textwidth}
  \centering
  \includegraphics[width=1\linewidth]{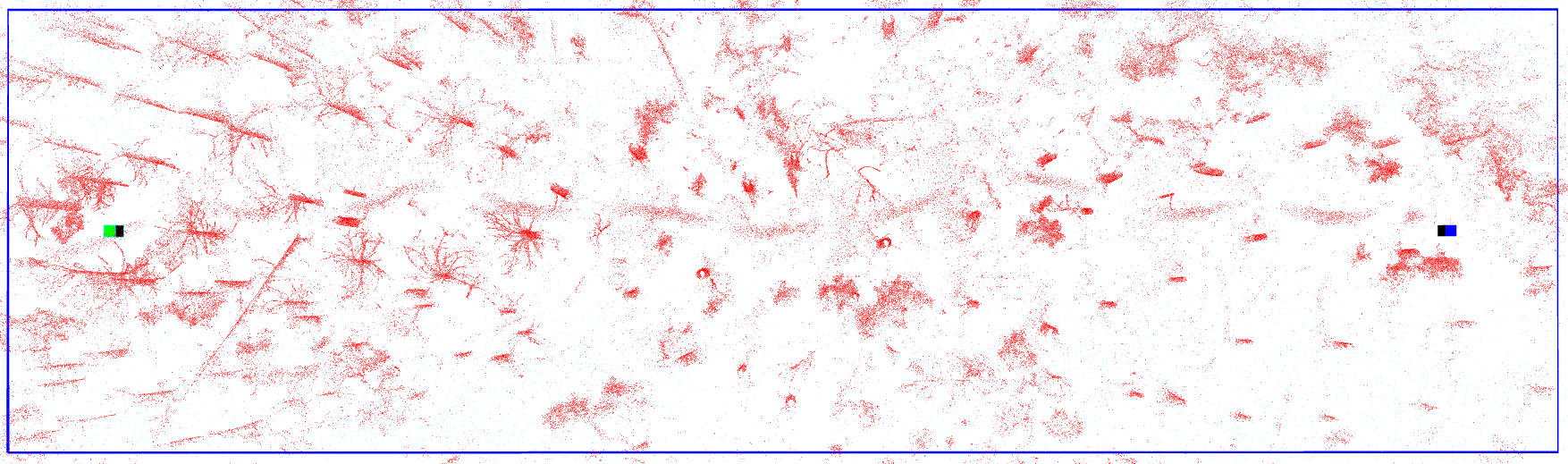}
  \caption{}
\end{subfigure}%

\begin{subfigure}{0.48\textwidth}
  \centering
  \includegraphics[width=1\linewidth]{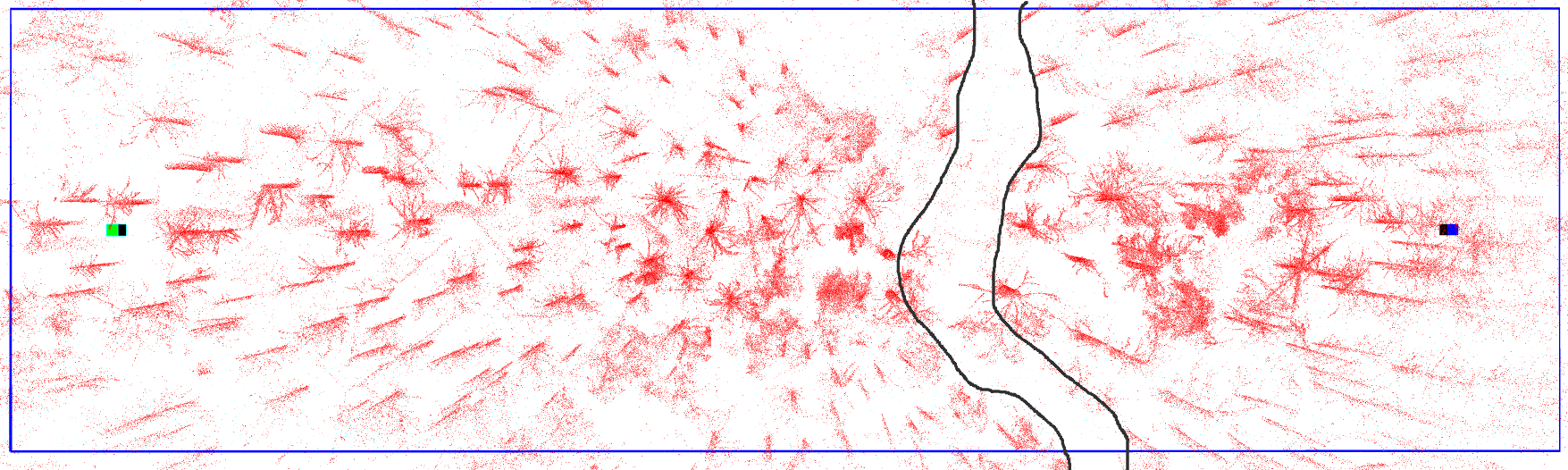}
  \caption{}
  \label{subfig:paloheinaPointCloudDif}
\end{subfigure}
\caption{Overhead view of the point cloud of the medium forest plot (a) and the difficult forest plot (b). The green cube depicts the starting position (0, 0, 1) and the blue cube depicts the goal position (60, 0, 1), both given in meters. The blue border marks the edges of the area where the density of the forest was calculated. The exact location of the goal position varied between flights due to slight variations in the initial heading between the flights. The black lines in (b) depict the edges of a trail, after which the denser end of the forest section starts.}
\label{fig:paloheina_pointcloud}
\end{figure}

\begin{figure}[tb]
\begin{subfigure}{0.48\textwidth}
  \centering
  \includegraphics[width=1\linewidth]{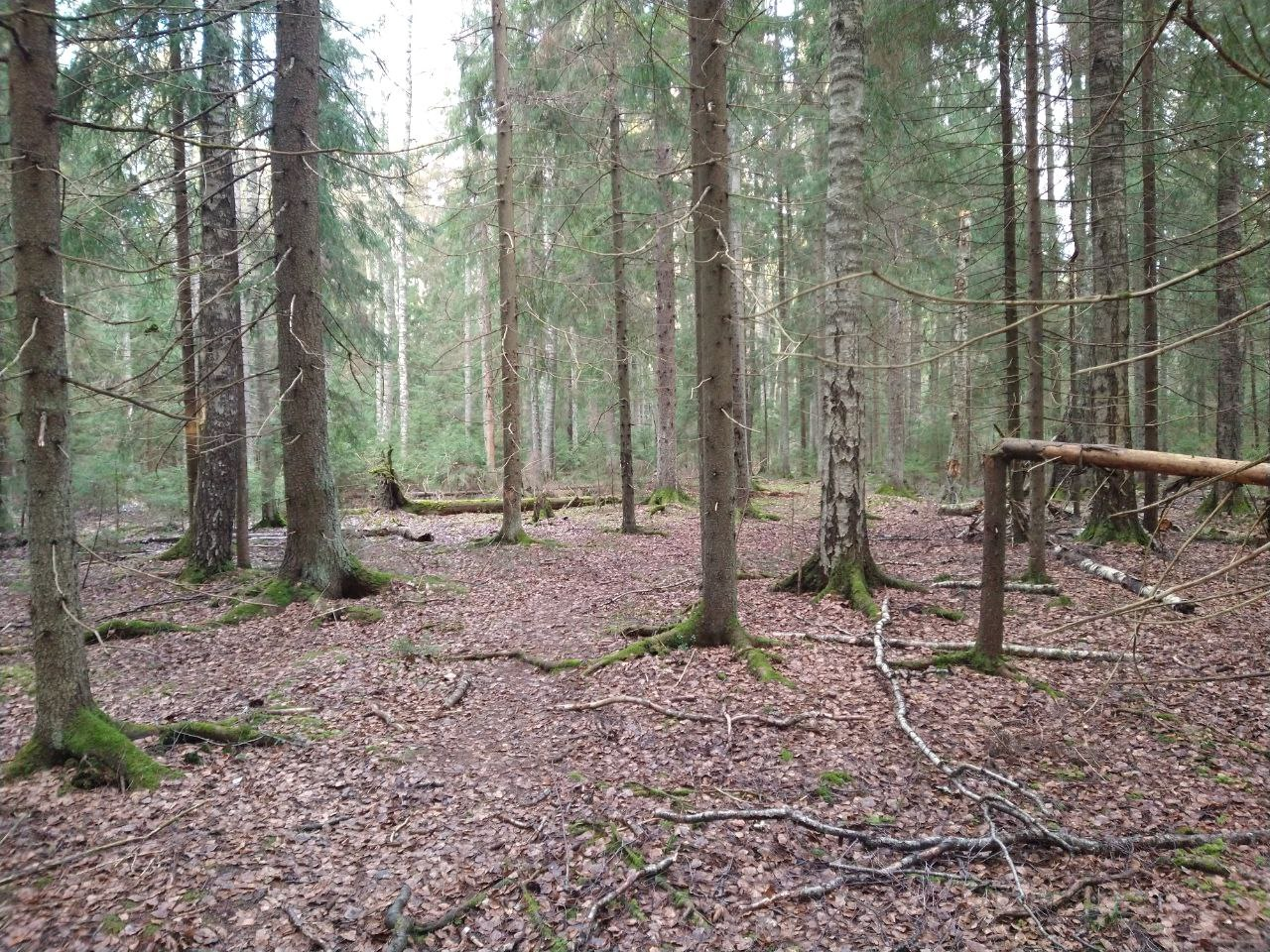}
  \caption{}
\end{subfigure}%

\begin{subfigure}{0.48\textwidth}
  \centering
  \includegraphics[width=1\linewidth]{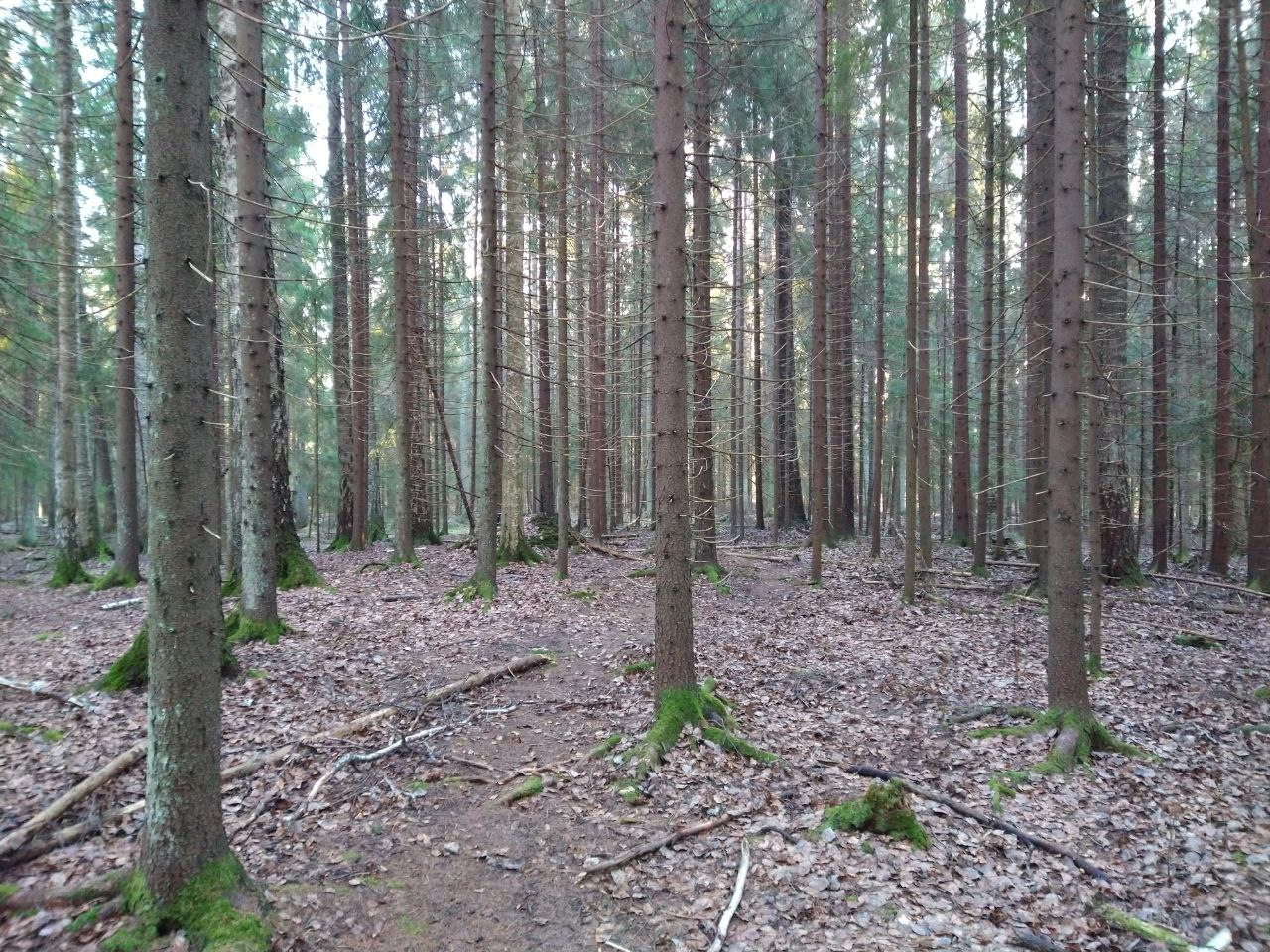}
  \caption{}
\end{subfigure}
\caption{View toward the goal from the starting position in the medium forest plot (a) and the difficult forest plot (b).}
\label{fig:paloheina_start}
\end{figure}

The performance of both the original and optimized systems was evaluated in two boreal forest areas located in Palohein\"{a}, Helsinki, Finland (60°15’28.4"N 24°55’19.9"E). Based on the complexity of the forests, the test plots were categorized as "medium" and "difficult" within a three-level categorization system. The categorization followed the forest density criteria presented by Liang et al. \cite{liang2019forest}, who divided boreal forest plots into three categories based on tree density and understory vegetation. Easy forests have a density of less than 700 trees/ha and minimal understory vegetation, medium forests have around 1000 trees/ha with sparse understory vegetation, and difficult forests have approximately 2000 trees/ha and dense understory vegetation. In this study, the medium and difficult plots had approximate tree densities of 1040 trees/ha and 2220 trees/ha, respectively. Overhead view of the point cloud acquired from both forests is presented in \autoref{fig:paloheina_pointcloud}. The density of both forests was approximated by manually calculating the number of trees from the point cloud inside the area depicted by the blue border in \autoref{fig:paloheina_pointcloud} and dividing that number by the area inside the blue border. The main tree species in both plots was Norway spruce (Picea abies). The amount of understory vegetation was low, but the spruces in the test areas had a high number of low-hanging dry branches, increasing the complexity and difficulty of the environment. Views from the starting locations in both forest plots are shown in \autoref{fig:paloheina_start}.

The performance of both the original and optimized systems was evaluated through flight experiments in two forest plots at target flight velocities of 1 m/s and 2 m/s. All flights followed a straight-line trajectory from the starting point to a goal point approximately 60 m ahead. At a target speed of 1 m/s, 15 flights per system were conducted in each plot, resulting in a total of 60 flights (30 per system). At a target speed of 2 m/s, three flights were conducted with the original system in the medium complexity forest, after which the experiments were aborted due to all flights failing. In contrast, the optimized system completed a full series of 30 flights at a target flight speed of 2 m/s, comprising 15 flights in the medium forest plot and 15 flights in the difficult forest plot.

For flights conducted in the medium forest plot, the goal point was located in free space, except for the first two flights performed with the original system. In contrast, for flights conducted in the difficult forest plot, the goal area was situated in a highly cluttered environment. Due to slight variations in the starting position and orientation of the quadrotor, no sufficiently large obstacle-free area could be identified. Consequently, for many flights in the difficult forest, the goal point was located within an obstacle.

Wind conditions were calm during all test flights. However, the condition of leaves on the ground and the number of leaves thrust upward by the quadrotor varied between flights (\autoref{tab:leafAmount}). 

\begin{table*}
\caption{The condition and amount of leaves on the ground, and how often the leaves would be thrust from the ground by the propellers of the quadrotor.}
\label{tab:leafAmount}
    \begin{tabularx}{\textwidth}{@{} p{6.5cm} *{2}{X} @{}}
    \toprule
    Algorithm, forest difficulty and target speed&
    Condition of leaves&
    Leaves thrust alongside  the quadrotor \\
    \midrule
    \textbf{Original system} & &\\
    Medium forest, 1 m/s & Lot of wet leaves & Often \\
    \rowcolor{lavender}
    Medium forest, 2 m/s  & Lot of wet leaves & Occasionally  \\
    Difficult forest 2 m/s & Lot of frozen leaves & Rarely \\
    \addlinespace
    \textbf{Optimized system} & &\\
    Medium forest 1 m/s & Moist leaves & Occasionally \\
    \rowcolor{lavender}
    Medium forest, 2 m/s (flights 1-8) & Moist leaves & Rarely \\
    Medium forest, 2 m/s (flights 9-15) & Dry leaves & Often \\
    \rowcolor{lavender}
    Difficult forest 1 m/s & Moist leaves & Occasionally \\
    Difficult forest 2 m/s & Dry leaves & Often \\
    \bottomrule
    \end{tabularx}
\end{table*}

The parameters of IPC were set as follows: The obstacle inflation of A*
and the SFC shrinking parameter was set to 0.4~m. The horizon length and time step of MPC were set to 15 and 0.1~s, respectively. The maximum speed, acceleration, and jerk were set for all directions to 10 m/s, 20 $\text{m/s}^2$, and 50 $\text{m/s}^3$, respectively, except for the downward acceleration, which was set to 9.5 $\text{m/s}^2$. The MPC problem weight parameters of \eqref{eq:mpcProb} were set as follows: $\mathbf{R}_u = \text{diag}(0,0,0)$, $\mathbf{R}_p = \text{diag}(2500,2500,2500)$, $\mathbf{R}_v = \text{diag}(0,0,0)$, $\mathbf{R}_a = \text{diag}(0,0,0)$, $\mathbf{R}_{p, N} = \text{diag}(3500,3500,3500)$, $\mathbf{R}_{v, N} = \text{diag}(200,200,200)$, $\mathbf{R}_{a, N} = \text{diag}(200,200,200)$, and $\mathbf{R}_c = \text{diag}(1.0,1.0,1.0)$. The forgetting threshold was set to 30 s with the original system and to 3 s with the optimized system. During preliminary testing with the original system, it was observed that increasing the forgetting threshold slightly reduced the vertical oscillations. In contrast, the optimized system performed better with a lower forgetting threshold. Additionally, for the optimized system, the weight parameter $w$ and the radius $d_{\text{follow}}$ for the old path following heuristics in \eqref{eq:AstarFollowHeuristic} were set to 150 and 5 m, respectively.

\subsection{PERFORMANCE ASSESSMENT}
\label{subsec:performanceAssessment}

The performance of both systems was evaluated using several metrics: success rate, flight quality, flight time, point-to-point average speed, average flying speed, and location accuracy. 

A flight was considered successful if the quadrotor either reached the target goal point or determined, when in close proximity, that the goal point was unreachable. 

Flight quality was assessed visually by identifying undesired behaviors, such as the number of emergency stops, the amount of unnecessary zigzagging, the number of collisions with tree trunks, branches, and general flight stability. 

The average flying speed was calculated as the average of the smoothed LTA-OM approximated speed over the entire flight. The average flying speed measured how close to the assigned target flight speed the quadrotor flew. The point-to-point average speed was computed by dividing the straight line distance from the start to the actual terminal point by the flight time of the mission, providing an estimate of the time required to traverse that distance. 

To quantify the time taken flying indirect routes, the following equation is used:
\begin{equation}
\label{eq:theoreticalShortestFlight}
t_{\text{extra}} = t_\text{true} - \frac{d}{v_\text{true}},
\end{equation}
where $t_{\text{extra}}$ is the time spent flying indirect routes during the flight, $t_\text{true}$ is the flight mission completion time, $d$ is the true flight distance, and $v_\text{true}$ is the average flying speed. If $t_{\text{extra}}$ is 0 s, then the flight would follow a perfectly straight line from start to finish. Therefore, when obstacles are present 
between the start and end points, achieving $t_{\text{extra}}$ of exactly 0 seconds is not possible. However, for an effective system, $t_{\text{extra}}$ should remain low.

The location accuracy of LTA-OM has been shown to be sufficient for data collection flights in forests. Zou et al. \cite{Zou20242455} compared the location accuracy of LTA-OM against four other lidar-based SLAM algorithms, FAST-LIO-SLAM \cite{FAST-LIO-SC}, SC-LIO-SAM \cite{SC-LIO-SAM}, SC-LeGO-LOAM \cite{SC-LeGO-LOAM} (all proposed by Kim), and LiLi-OM proposed by Li et al. \cite{liliom} using the Mulran \cite{gskim-2020-mulran} and NCLT \cite{carlevaris2016university} datasets. Across nine data sequences from both datasets, LTA-OM achieved the lowest average location root-mean-square error (RMSE), of 4.83 m for the Mulran dataset and 1.56 m for the NCLT dataset. As both datasets were collected in urban environments and with higher-quality lidar than the Livox Mid-360, these results cannot be directly transferred to forest environments. In a study by Karhunen et al. \cite{isprs-archives-XLVIII-2-W11-2025-153-2025}, the performance of LTA-OM was evaluated in a forest environment using the Livox Mid-360 lidar. Two manual piloted flights with distances of 150 m and 420 m were conducted in a forest in Palohein\"{a}, Helsinki, and the endpoint drift was measured at 0.19 m  and 0.08 m, respectively. Based on the documented performance, the localization performance of the algorithms was considered sufficient. Therefore, this study conducted a simple validation of location accuracy by comparing LTA-OM height estimates with actual height differences determined through visual analysis.

\section{RESULTS} \label{sec:results}

\subsection{DESCRIPTION OF REASONS OF FAILURES}
\label{subsec:descriptionFailures}

The causes of flight failures were divided into four categories: collisions caused by leaf clouds being thrust from the ground around the quadrotor, motor shutdowns due to NaNs generated during the SFC generation process, collisions resulting from unstable flight, and collisions with trees. 

The cause of failures due to clouds of leaves being thrust around the quadrotor was that the leaves caused the quadrotor to mistakenly detect itself as being inside an obstacle. Since A* cannot perform reference path planning when the starting location is within an obstacle, the starting point of the search was moved to the closest free voxel in the obstacle map. This discrepancy between the actual location of the quadrotor and the generated reference path sometimes led to very aggressive control inputs or to an MPC solver failure. Recovery from the solver failure while adhering to the MPC constraints could be difficult. Such extended MPC solver failures could result in collisions with nearby obstacles. However, in many cases, clouds of leaves being thrust from the ground did not lead to a failure. 

The cause of failures due to NaNs generated during the SFC generation process was that the corners of the SFC polyhedron were defined as NaN values. If these NaNs were not properly handled, the invalid SFC corner NaN values were forwarded to the MPC problem solver, which in turn produced NaN-valued control commands. When forwarded to the PX4 FCU, these commands, in response, turned the motors off. 

\begin{figure*}[tb]
\begin{subfigure}{.33\linewidth}
  \centering
  \includegraphics[width=.99\linewidth]{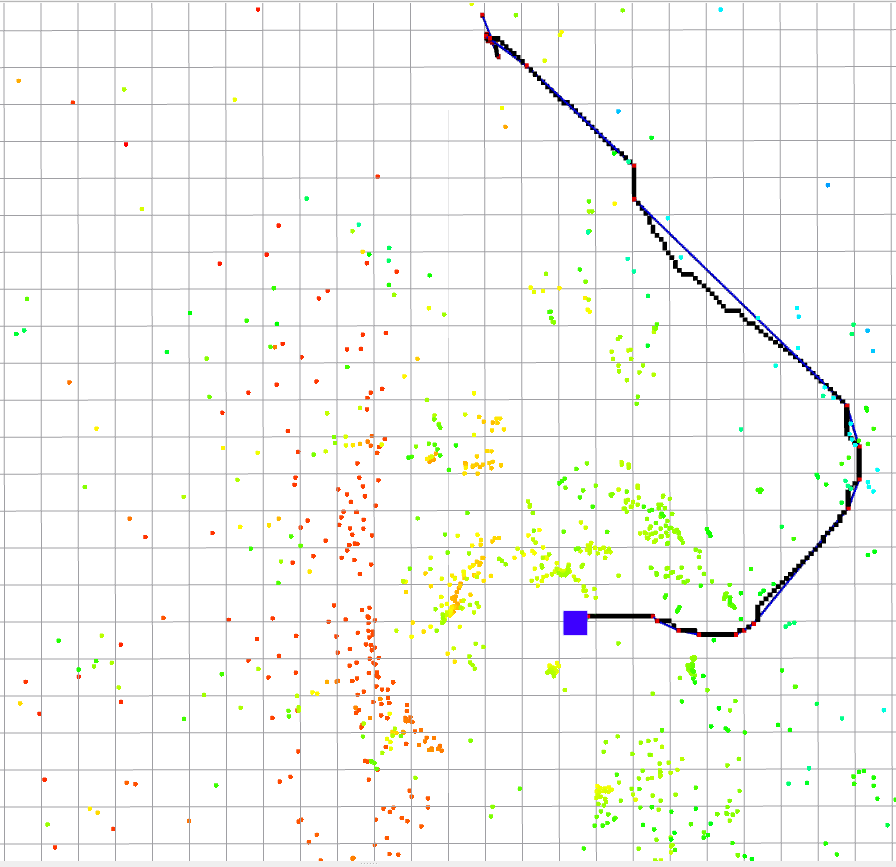}
  \caption{}
  \label{fig:Astar_1}
\end{subfigure}%
\begin{subfigure}{.33\linewidth}
  \centering
  \includegraphics[width=.99\linewidth]{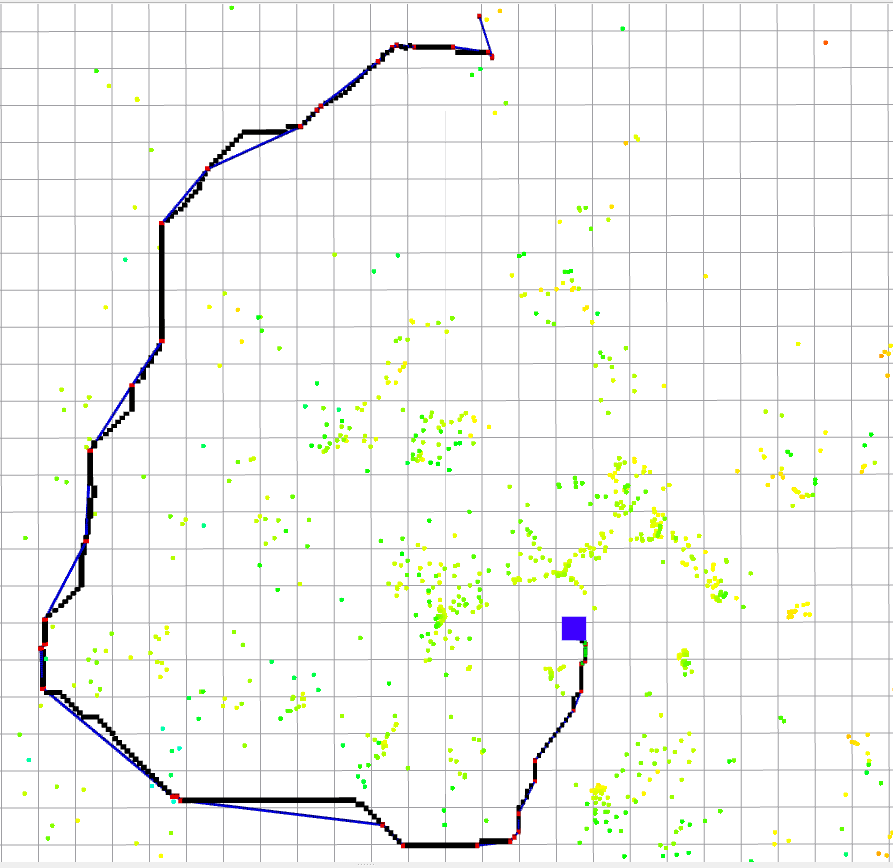}
  \caption{}
  \label{fig:Astar_2}
\end{subfigure}
\begin{subfigure}{.33\linewidth}
  \centering
  \includegraphics[width=.99\linewidth]{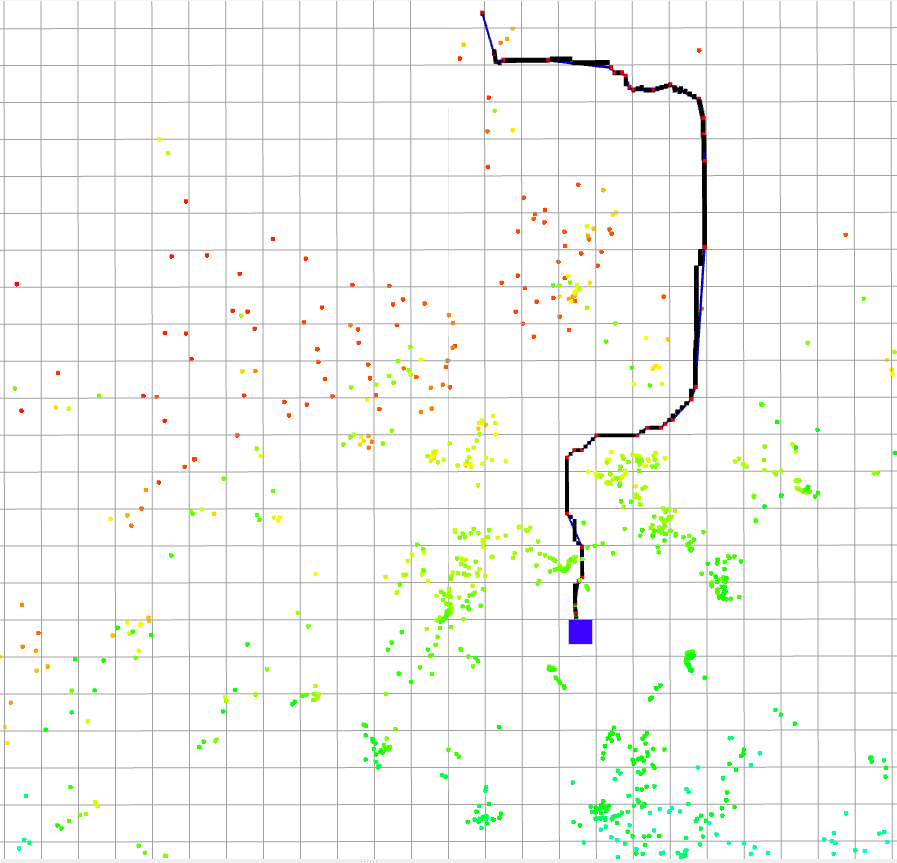}
  \caption{}
  \label{fig:Astar_3}
\end{subfigure}
\caption{An example of volatile A* reference path planning from flight 10 of the flights done with the original system in the difficult forest. The blue box depicts the current position of the quadrotor, the black line depicts the optimal planned A* path, and the blue line depicts the output flight path. The side length of each cell of the grid is 1 meter. The time interval between the first picture (a) and the last picture (c) was approximately 1.2 seconds.}
\label{fig:volatileAstar}
\end{figure*}

Failures due to unstable flight were caused by prolonged MPC solver failures that ultimately led to collisions. Such failures occurred when the velocity and location of the quadrotor were such that no feasible control input could keep it within the SFCs. In these situations, the system continued to apply the last valid set of control commands obtained before the MPC solver failure. If the solver failure persisted, these outdated control inputs steered the quadrotor to collide with nearby obstacles. Several root causes contributed to prolonged MPC solver failures, with volatile A* path planning being the most significant. In the original system, each re-planning event caused A* to search for the shortest path towards the goal without considering the previously planned path. As a result, successive reference paths could begin in entirely different directions. An example of this behavior is illustrated in \autoref{fig:volatileAstar}. Beyond causing prolonged MPC solver failures, particularly in highly cluttered areas, this behavior significantly increased mission completion times, as the quadrotor maneuvered around the same position for extended periods.

Failures due to collisions with trees occurred when the quadrotor followed a reference path that passed extremely close to a tree. This behavior resulted from the optimality property of A*, which generates paths that go around obstacles along the nearest free voxels. However, in certain situations, the system was unable to track the reference path with sufficient accuracy, causing the quadrotor to enter an obstacle in the obstacle map without physically colliding with the tree. This was possible because the obstacle inflation parameter was set larger than the actual radius of the quadrotor. Once the quadrotor was considered to be inside an obstacle, the MPC problem could no longer be solved, as the initial state was located outside the feasible SFC. In these situations, the system continued to apply the control commands from the last successful MPC solution, which ultimately directed the quadrotor to collide with the tree. The inability to track the reference path tightly was caused by various factors, including unstable flight behavior and wind disturbances.

\autoref{tab:flightOverview} provides an overview of the results of the flights, and a more detailed overview of the individual flights is presented in the Appendix.  A summary of the number of failures by cause for both system versions across all flight velocities and forest difficulty levels is shown in \autoref{tab:failuersNumbers}. Video recordings of several flights are provided in the supplementary materials.

\begin{table*}
    \caption{Overview of all flight experiments. Point-to-point average speed, average flying speed, and $t_{\text{extra}}$ were measured for only successful flights. An "x" indicates that no data was collected.}
    \label{tab:flightOverview}
    \begin{tabularx}{\textwidth}{@{} p{4.5cm} *{6}{C} @{}}
    \toprule
    Algorithm, forest difficulty and target speed&
    Success rate&
    Collisions (which lead to a failure)&
    Aggressive leaf dodges (which lead to a failure)&
    grand mean point-to-point speed&
    grand mean flying speed &
    $t_{\text{extra}}$ from the average speeds\\
    \midrule
    \textbf{Original system} & & & & & &\\ 
    Medium forest, 1 m/s & 10/15 (67 \%) & 12 (1) & 19 (2) & 0.68 m/s & 0.76 m/s & 12.5 s\\ 
    \rowcolor{lavender}
    Medium forest, 2 m/s & 0/3 (0 \%) & 1 (1) & 4 (0) & x & x & x\\ 
    Difficult forest, 1 m/s & 6/15 (40 \%) & 22 (4) & 1 (0) & 0.44 m/s & 0.70 m/s & 60.4 s\\
    \addlinespace
    \textbf{Optimized system} & & & &\\
    Medium forest, 1 m/s & 12/15 (80 \%) & 2 (0) & 14 (3) & 0.76 m/s & 0.80 m/s & 3.2 s\\ 
    \rowcolor{lavender}
    Medium forest, 2 m/s & 12/15 (80 \%) & 9 (2) & 16 (2) & 1.35 m/s & 1.48 m/s & 4.0 s\\
    Difficult forest, 1 m/s & 15/15 (100 \%) & 2 (0) & 9 (0) & 0.66 m/s & 0.74 m/s & 9.7 s\\
    \rowcolor{lavender}
    Difficult forest, 2 m/s & 5/15 (33 \%) & 13 (4) & 19 (5) & 1.06 m/s & 1.32 m/s & 12.0 s\\
    \bottomrule
    \end{tabularx}
\end{table*}

\begin{table*}
    \caption{Overview of the failure reasons for all flight experiments. *With the original system in the medium forest, with a target flight speed of 2 m/s, only three flights were performed.}
    \label{tab:failuersNumbers}
    \begin{tabularx}{\textwidth}{@{} p{4.5cm} *{5}{C} @{}}
    \toprule
    Algorithm, forest difficulty and target speed&
    Hit a tree&
    Cloud of leaves&
    NaN SFC&
    Unstable flying&
    Total\\
    \midrule
    \textbf{Original system} & & & &\\ 
    Medium forest, 1 m/s & 1 & 2 & 1 & 1 & 5\\ 
    \rowcolor{lavender}
    Medium forest, 2 m/s* & 0 & 0 & 0 & 3 & 3*\\
    Difficult forest, 1 m/s & 3 & 0 & 4 & 2 & 9\\
    \addlinespace
    \textbf{Optimized system} & & & &\\ 
    Medium forest, 1 m/s & 0 & 3 & 0 & 0 & 3\\ 
    \rowcolor{lavender}
    Medium forest, 2 m/s & 0 & 2 & 0 & 1 & 3\\
    Difficult forest, 1 m/s & 0 & 0 & 0 & 0 & 0\\
    \rowcolor{lavender}
    Difficult forest, 2 m/s & 2 & 5 & 0 & 3 & 10\\ 
    \bottomrule
    \end{tabularx}
\end{table*}

\subsection{FLIGHT TESTS WITH THE ORIGINAL SYSTEM}
\label{subsec:orgFlightTest}

\subsubsection{FLIGHTS IN THE MEDIUM FOREST}

\begin{figure*}[tb]
    \includegraphics[width=.99\linewidth]{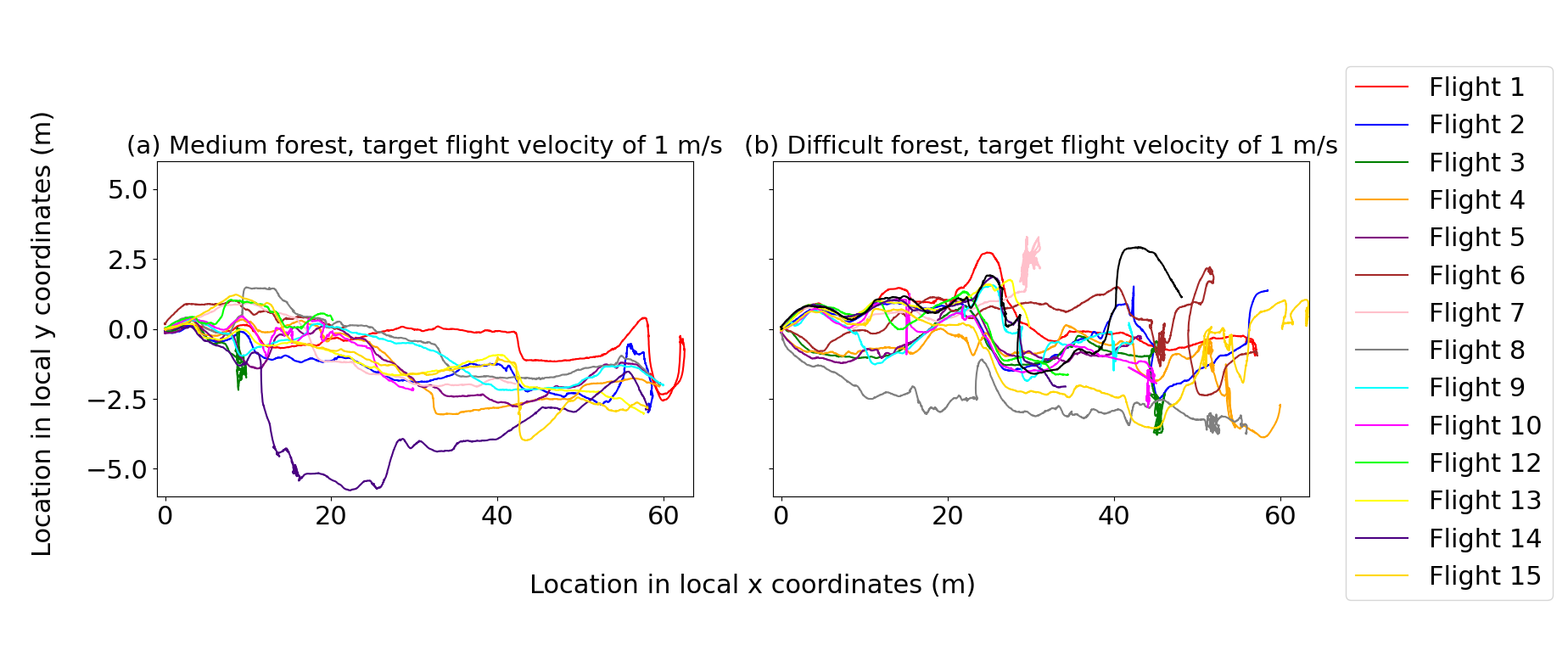} 
    \caption{The LTA-OM approximated flight paths of all flights flown with a target flight speed of 1 m/s in the medium forest (a) and in the difficult forest (b) with the original system. The flight paths are given in the local coordinates of individual flights, which means that the same location on the graph does not necessarily correspond to the same location in the test area. For the flights done in the medium forest (a), the rosbag recording of flight 11 was lost and is omitted from the graph. For the flights done in the difficult forest (b), the rosbag recording of flight 7 was cut short, but the terminal point of that flight is very close to the one depicted in the graph.}
    \label{fig:trailOrgFlights}
\end{figure*}

A total of 10/15 flights were successful with the original system in the medium forest with a target flight speed of 1 m/s (\cref{tab:flightOverview,tab:paloheinaOrgMed}). Of the five failed flights, two failures were caused by clouds of leaves being thrust from the ground around the quadrotor, one resulted from a collision with a tree, one was due to NaNs generated during the SFC generation process, and one was caused by unstable flying. During the flights, clouds of leaves caused the quadrotor to aggressively dodge the leaves a total of 19 times, two of which resulted in failures. Overall, flight behavior in the medium forest was mostly stable, with occasional aggressive maneuvers when avoiding leaves or nearby obstacles. This behavior can be seen in \autoref{fig:trailOrgFlights} as a relatively low number of zigzags and aggressive turns in the flight trajectories. In seven flights, a total of 12 collisions with branches and other foliage were observed, one of which led to a failure. In the remaining cases, the quadrotor was able to recover and continue the flight after the collision.

\begin{figure}[tb]
\begin{subfigure}{0.24\textwidth}
  \centering
  \includegraphics[width=.99\linewidth]{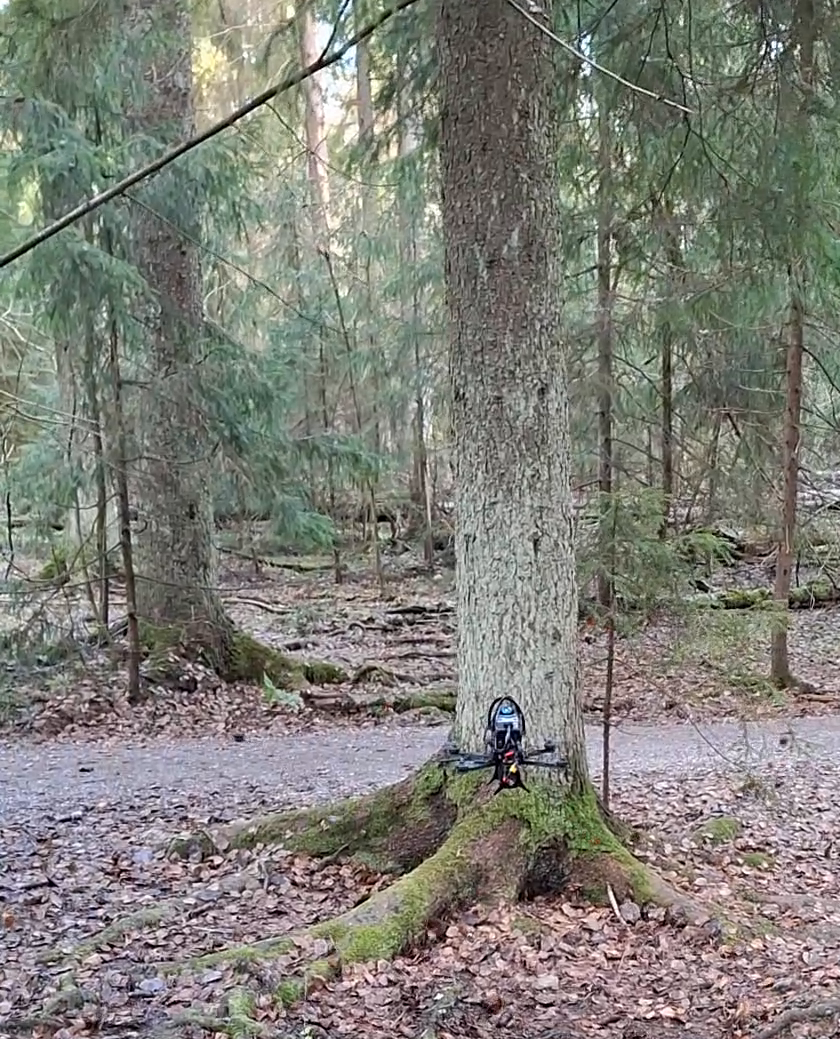}
  \caption{}
\end{subfigure}%
\begin{subfigure}{0.24\textwidth}
  \centering
  \includegraphics[width=.99\linewidth]{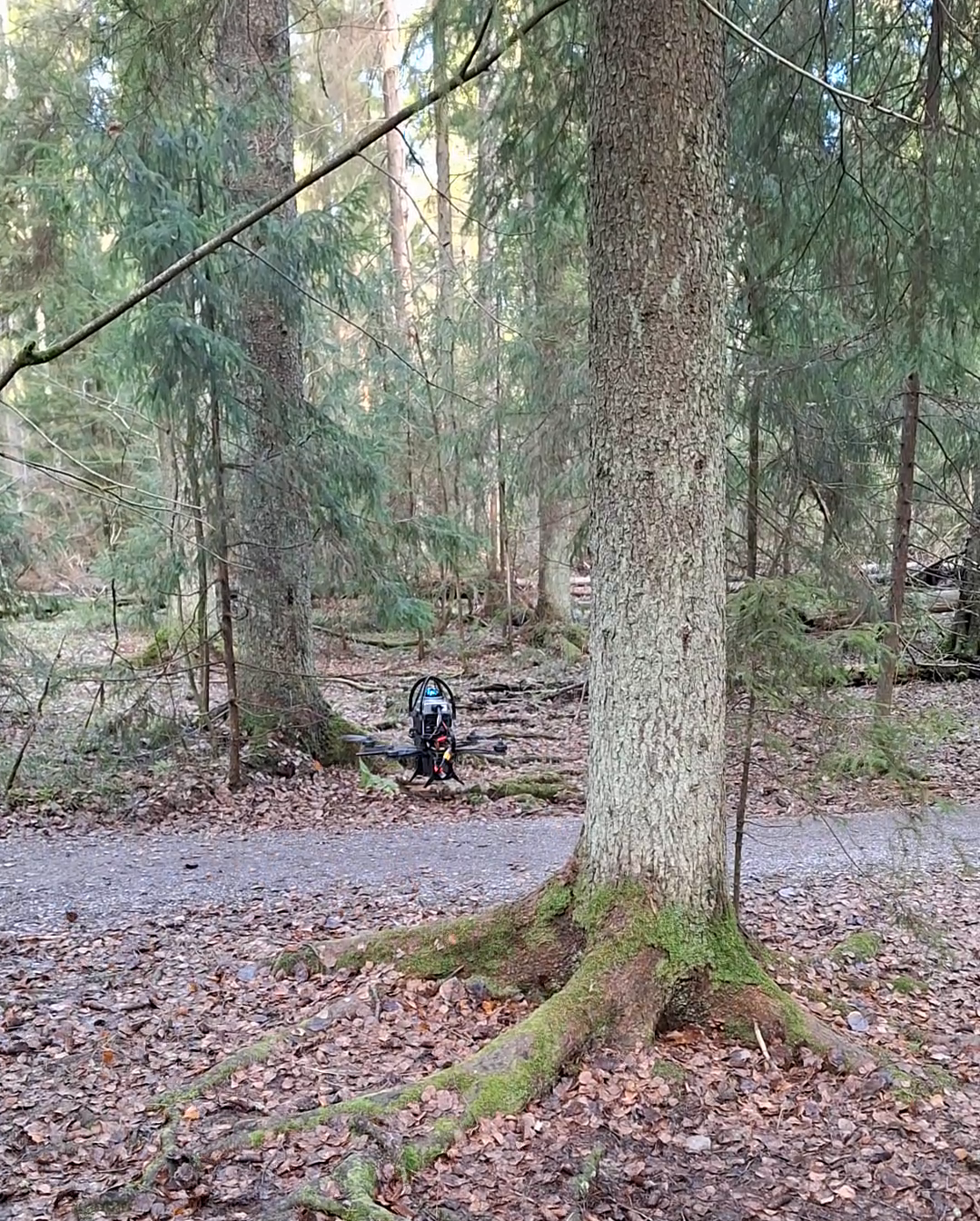}
  \caption{}
\end{subfigure}
\caption{The terminal locations of flights eight (a) and nine (b) performed in the medium forest with the original system. The height difference of approximately 30 cm is clearly visible between the two flights. The difference in the xy-plane is due to a slightly different orientation of the starting location.}
\label{fig:LTAOMheightEst}
\end{figure}

Significant differences were also observed in the height estimates produced by LTA-OM. Although no ground truth position data were collected, visual inspection revealed notable discrepancies in altitude estimates between flights. \autoref{fig:LTAOMheightEst} illustrates the terminal location of flights eight and nine. While the LTA-OM estimated terminal heights differed only 2 cm (1.00 m versus 0.98 m), visual evidence suggests that the actual height difference was approximately 30 cm.

The average flying speed of the quadrotor was consistently below the target flight speed of 1 m/s, ranging from 0.60 m/s to 0.81 m/s, with a grand mean of 0.76 m/s (\cref{tab:flightOverview,tab:paloheinaOrgMed}). The point-to-point average speed was lower still, ranging from 0.43 m/s to 0.77 m/s, with a grand mean of 0.68 m/s. Consequently, the 60-meter flight missions required an average of 89.2 s to complete. The longest mission lasted 133.8 s, while the shortest mission was completed in 75.2 s. Most flights were completed in approximately 80 s, although three flights required significantly longer times. The average $t_{\text{extra}}$ per flight remained relatively low, at 12.5 s.

All flights with a target flight speed of 2 m/s  failed before reaching 20 m due to volatile A* replanning, which led to prolonged MPC solver failures (\cref{tab:flightOverview,tab:paloheinaOrgMed2ms}). After three flights, it was deemed highly unlikely that the original system could perform any successful flights at this speed. To avoid hardware damage resulting from collisions, no further flights were conducted with the original system at a target flight speed of 2 m/s.

\subsubsection{FLIGHTS IN THE DIFFICULT FOREST}

With the original system, six of the 15 flights in the difficult forest with a target flight speed of 1 m/s were successful (\cref{tab:flightOverview,tab:paloheinaOrgDif}). Of the nine failed flights, three were due to collisions with trees, four were caused by NaNs generated during the SFC generation process, and two resulted from unstable flying. None of the failures were caused by clouds of leaves being thrust around the quadrotor. In fact, only a single instance occurred where the system executed an aggressive leaf-dodging maneuver. In general, flying in the difficult forest was unstable, characterized by frequent aggressive maneuvers, collisions with small branches (22 in total, four of which led to a failure), and repetitive flying in place, illustrated in \autoref{fig:trailOrgFlights} by the high number of zigzags and sharp turns in flight paths. The most pronounced example of repetitive in-place flying occurred in the seventh flight, where the quadrotor flew within a dense group of trees for approximately six minutes. 

The average flying speed was substantially below the target of 1 m/s, ranging from 0.64 m/s to 0.76 m/s, with a grand mean of 0.70 m/s (\cref{tab:flightOverview,tab:paloheinaOrgDif}). Moreover, the point-to-point average speed was significantly lower across all flights, ranging from 0.31 m/s to, and the highest was 0.59 m/s, with a grand mean of only 0.44 m/s. Hence, on average, the 60-meter flight missions took 144.3 seconds to complete. The variation in flight mission completion times was large, with the shortest taking 94.8 seconds and the longest taking 194.7 seconds. Most successful flights, namely 4 out of 6, were over 140 seconds long, due to long periods of in-place flying. For those flights, $t_{\text{extra}}$ was between 59 and 99 seconds. On average $t_{\text{extra}}$ was 60.4 seconds

\subsection{FLIGHT TESTS WITH THE OPTIMIZED SYSTEM}

\subsubsection{FLIGHTS IN THE MEDIUM FOREST}

\begin{figure*}[tb]
    \includegraphics[width=.99\linewidth]{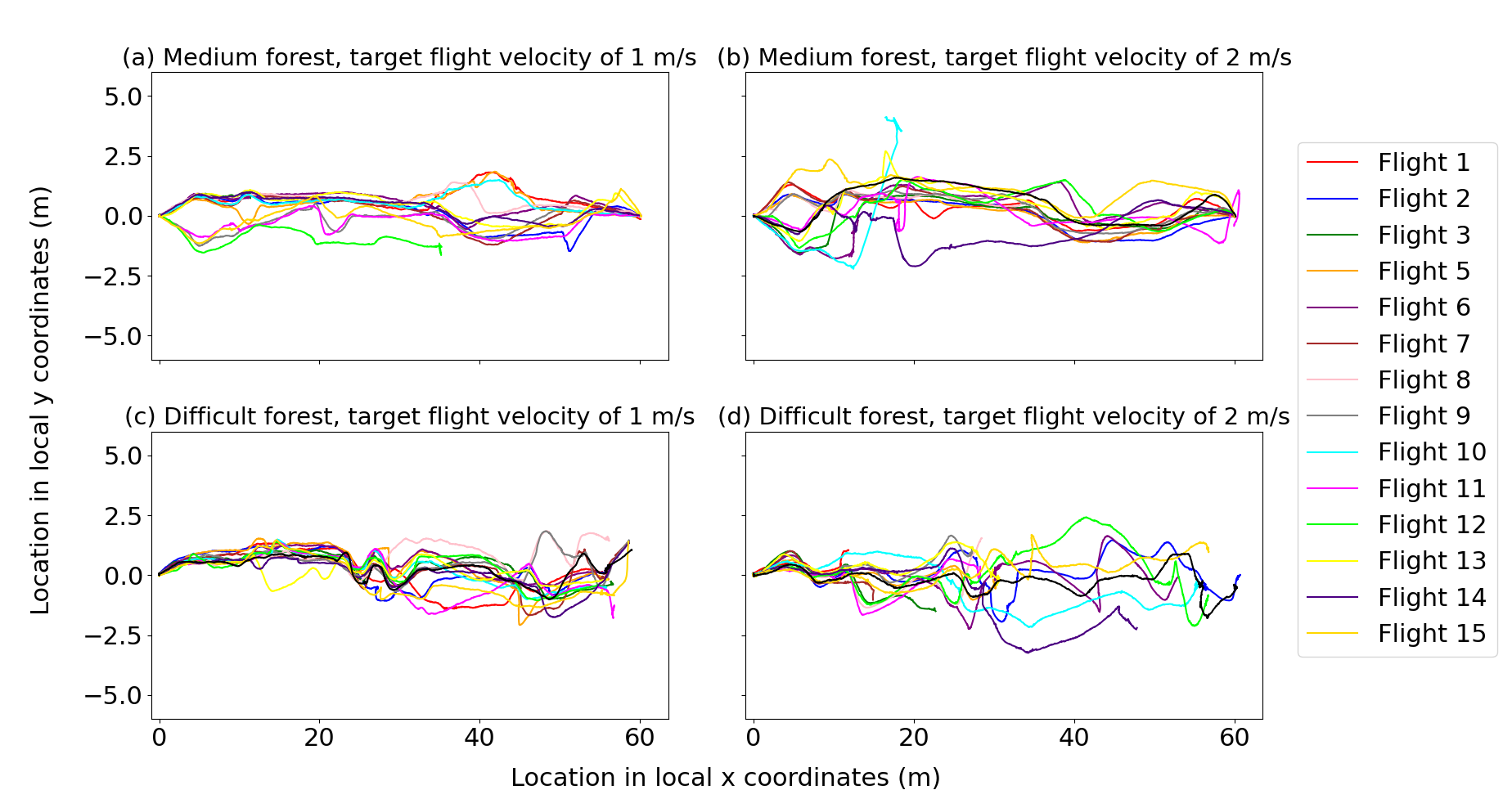}
    \caption{The LTA-OM approximated flight paths of all flights flown with the optimized system. (a) and (b) show the flight paths flown in the medium forest, and (c) and (d) in the difficult forest. The flight paths are given in the local coordinates of individual flights, which means that the same location on the graph does not necessarily correspond to the same location in the test area. Flights (a) and (c) were flown with a target flight speed of 1 m/s, while (b) and (d) were flown at 2 m/s.}
    \label{fig:trailOptFlights}
\end{figure*}

A total of 12 out of 15 flights were successful in the medium forest with a target flight speed of 1 m/s (\cref{tab:flightOverview,tab:paloheinaOptMed}). In all failure cases, a cloud of leaves was thrust around the quadrotor, which led to a collision with the ground. The cloud of leaves also affected flight behavior in 10 of the successful flights, during which a total of 14 aggressive leaf dodges were executed. In general, the flights were smooth, with few aggressive maneuvers, except when avoiding leaves or navigating close to obstacles. This behavior can be seen in \autoref{fig:trailOptFlights}, which shows almost no zigzagging or aggressive turns in the flight paths. The generally smooth flight behavior also resulted in only two collisions with obstacles during the flights.


The difference between the average flying speed and the average point-to-point speed was small (\cref{tab:flightOverview,tab:paloheinaOptMed}). The average flying speed was lower than the target flight speed of 1 m/s, ranging from 0.70 m/s to 0.84 m/s, with a grand mean of 0.80 m/s. The point-to-point average speed ranged from 0.65 to 0.79 with a grand mean of 0.76 m/s. Consequently, the 60-meter flight missions lasted 78.7 s on average The longest flight durations were 93.1 and 81.7 s, while the shortest was 75.6 s. Due to linear and non-zigzag routes taken, $t_{\text{extra}}$ remained small; $t_{\text{extra}}$ varying between 0.7 and 6.7 s.

The flights in the medium difficulty forest at a target flight speed of 2 m/s  were flown over three separate days (\autoref{tab:paloheinaOptMed2ms}).
The first eight flights were conducted on the first day, the ninth flight on the second, and the remaining six flights on the third day. During the first eight flights, there were moist leaves on the ground, which were rarely thrust upward from the ground. In contrast, during the last seven flights, the ground was covered with dry leaves that were frequently thrust upward from the ground.

With the optimized system, 12 of 15 flights with the target speed of 2 m/s were successful (\cref{tab:flightOverview,tab:paloheinaOptMed2ms}). Two of the failed flights were caused by clouds of leaves, and one was due to unstable flying caused by a prolonged MPC solver failure. Throughout the experiments, the system was frequently required to dodge leaves. During the first seven flights, aggressive leaf-avoidance maneuvers occurred three times, whereas during the last eight flights, 13 such maneuvers were observed. In general, flights with a target speed of 2 m/s were less stable than those with 1 m/s, showing more frequent and aggressive leaf-avoidance maneuvers, which can be as sharp turns in the flight paths shown in \autoref{fig:trailOptFlights}. The more unstable flying also led to more collisions with the surrounding obstacles. In total, there were nine collisions, two of which led to failure. 

The variation of the average flight speed was small (\cref{tab:flightOverview,tab:paloheinaOptMed2ms}). The average flight speed was lower than the target flight speed of 2 m/s, ranging from 1.23 to 1.58 m/s, with the grand mean of 1.48 m/s. The point-to-point average speed ranged between 1.14 and 1.51 m/s, with the grand mean of 1.35 m/s. Although the absolute difference between the average flight speed and the point-to-point average speed was larger than in the 1 m/s case, the higher absolute speed of the quadrotor resulted in low $t_{\text{extra}}$ values. $t_{\text{extra}}$ ranged from  1.2 to 9.5 s. Flight mission completion times ranged from 39.6 to 51.7 s with an average duration of 41.3 s. With the optimized system, emergency stop procedures were triggered when no path to the goal could be found within 0.1 s. For most flights, these emergency stops increased mission completion times by 0.5--3.0 seconds. However, during the 15th flight, an extended emergency stop occurred, after which the system concluded that no feasible path to the goal existed. Consequently, the goal command was reissued after 11 s.

With both target flight velocities, there were no visually observable differences in height estimation. Based on visual inspection of the quadrotor while hovering at the goal location, the quadrotor appeared to maintain a consistent altitude across all completed flights. 

\subsubsection{FLIGHTS IN THE DIFFICULT FOREST}

All 15 flights conducted in the difficult forest at a target speed of 1 m/s were successful (\cref{tab:flightOverview,tab:paloheinaOptDif}). The number of aggressive leaf-avoidance maneuvers was moderate, with a total of nine such dodges observed across all flights. Overall, flight behavior was relatively stable with only a few aggressive maneuvers, excluding emergency stops triggered when A* path planning exceeded the 0.1 s limit. The relatively stable behavior is evident in \autoref{fig:trailOptFlights}, where only a few aggressive turns appear in the flight trajectories. The stability of the flights was also reflected in a low number of collisions with obstacles, with only two collisions during the flights. 

Since the goal point was located in a very dense area, the quadrotor had challenges in finding a feasible route to the goal point. Concequently, in every flight the A* path search exceeded 0.1 s at least once near the goal, triggering the emergency stop procedure. In most flights, the emergency stop was activated 1-4 times; however, in two flights it was triggered nine and ten times, respectively. Furthermore, ten flights terminated during an emergency stop after the quadrotor determined that no feasible path to the goal existed. In the remaining five flights, the flight ended near the goal point, which was repositioned by the path planner module after the original goal was deemed to be located inside an obstacle.

The average flight speed was lower than the target speed of 1 m/s, ranging from 0.62 m/s to 0.79 m/s, with a grand mean of 0.74 m/s (\cref{tab:flightOverview,tab:paloheinaOptDif}). The point-to-point average speed varied between 0.51 m/s and 0.73 m/s, with a grand mean of 0.67 m/s. Flight durations ranged from 76.6 to 115.9 s with an average of 86.5 s. $t_{\text{extra}}$ remained small for most flights, ranging from 3.7 to 22.0 s. In the difficult forest, emergency stops triggered by the A* module exceeding the 0.1 s planning threshold lasted longer than in the medium forest. For most flights, the cumulative duration of emergency stop periods ranged from 0 to 6 s; however, in two flights (6 and 14), these periods extended to 26 and 22 s, respectively. 

In the difficult forest, five out of fifteen flights were successful with a target flight speed of 2 m/s (\cref{tab:flightOverview,tab:paloheinaOptDif2ms}). Out of the ten unsuccessful flights, five failed due to a cloud of leaves, three due to unstable flying, and two due to a collisions with trees. A large number of leaves were thrust up from the ground during the flights, requiring the system to aggressively dodge leaf clouds 19 times in total, five of which resulted in failure. As with the flights conducted at a target flight speed of 1 m/s, numerous emergency stops were triggered, particularly after the trail, due to the A* path path planner exceeding the allowable planning time. Each successful flight involved 1-8 emergency stops, with some occurring immediately  after takeoff, similar to the behavior observed in the medium forest at a target speed of 2 m/s. Three successful flights terminated after the system determined that no feasible path to the goal existed, while the remaining two ended near the goal position, which was repositioned by the path planner module. Overall, flying was more unstable than at a target speed of 1 m/s or in the medium forest. This is evident in \autoref{fig:trailOptFlights}, which shows a higher number of aggressive turns in the flight paths. The increased instability also led to a higher number of collisions with branches, with 13 collisions in total, four of which resulted in failure. 

The average flying speed was substantially lower than the target flight speed of 2 m/s, ranging from 1.06 to 1.56 m/s, with a grand mean of 1.32 m/s (\cref{tab:flightOverview,tab:paloheinaOptDif2ms}). The point-to-point average speed varied between 0.79 and 1.35 m/s, with a grand mean of 1.06 m/s. Flight mission durations ranged from 42.0 to 76.1 s, with an average of 56.8 s. $t_{\text{extra}}$ ranged from 5.7 to 19.3 s. Of the five successful flights, three spent only 1-3 s in emergency stop periods, whereas the remaining two (flights 2 and 15) spent 10 and 19 s in emergency stops, respectively.

\subsection{ANALYSIS AND COMPARISON OF ORIGINAL AND OPTIMIZED SYSTEMS}

The performance of the original system was poor, particularly in the difficult forest. With a success rate of only 40 \% and an average of nearly a full minute spent flying indirect and zigzag routes, the system proved unsuitable for flights in boreal forests. Although performance was better in the medium forest, reliability still needed improvement. Moreover, the original system failed to complete any flights at a target speed of 2 m/s, indicating that increasing flight speed to reduce mission completion times would be inadvisable. To address these limitations, the optimized system aimed to improve reliability, reduce mission completion times by minimizing $t_{\text{extra}}$, and increase average flight speed without compromising reliability. Additionally, the original system exhibited issues with height estimation in the medium forest. 

The reliability of the optimized system was substantially higher than that of the original system. At a target flight speed of 1 m/s, the success rate increased from 10/15 to 12/15 in the medium forest and from 6/15 to 15/15 successful flights in the difficult forest. This improvement was primarily due to the less volatile A* path planning of the enhanced reference path module, which stabilized flight and reduced aggressive maneuvers, particularly in cluttered areas. The improved stability is evident in the flight paths shown in \Cref{fig:trailOrgFlights,fig:trailOptFlights}, which exhibit minimal zigzagging and fewer sharp turns compared with the original system. The optimized system also experienced fewer collisions that led to failures; in the medium forest, collisions decreased from 12 to 2, and in the difficult forest, from 22 to 2, at a target flight speed of 1 m/s. Even at a target speed of 2 m/s, collisions were reduced compared with the original system, with 9 collisions in the medium forest and 13 in the difficult forest.

Another improvement was the handling of NaNs during the SFC generation process. Although the error handling was simple and occasionally resulted in extended emergency stop periods, it completely eliminated the failures caused by NaNs.

Clouds of leaves posed challenges for both systems. Indeed, all failures at a target flight speed of 1 m/s were caused by leaf clouds. The proportion of aggressive leaf dodges that led to a failure was similar for both systems at a target speed of 1 m/s (10\% for the original system and 13 \% for the optimized system). However, visual observation suggested that the optimized system dodged leaves less aggressively than the original system, resulting in fewer total aggressive maneuvers. Overall, effectively handling clouds of leaves rising from the ground remains a challenge, even for the optimized system.

Lastly, the optimized system resolved the height estimation issue observed in the original system. By incorporating gravity direction measurements, the end-point height of flights was maintained consistently.

The improved flight stability and reliability of the optimized system were particularly evident at the higher target flight speed of 2 m/s. With the original system, none of the flights were successfully completed, even in the medium forest. In contrast, the optimized system completed 12 out of 15 flights in the same conditions. 

There was no significant reduction in the reliability of the optimized system at a target flight speed of 1 m/s as forest difficulty increased. In fact, the success rate improved from 12 out of 15 flights in the medium forest to 15 out of 15 flights in the difficult forest. Based on these experimental results, the reliability of the system at low target flight velocities does not appear to be strongly affected by the density of the forest. 

The reliability of the optimized system decreased as the target flight speed increased from 1 m/s to 2 m/s in both forest densities, particularly in the difficult forest. In the medium forest, 12 of 15 flights were successful at both target flight velocities; however, higher flight speeds resulted in generally less stable flight behavior, leading to an increased number of collisions with branches that did not cause mission failures (2 at 1 m/s versus 7 at 2 m/s). Overall, the reduction in reliability in the medium forest was minor. In contrast, performance degradation in the difficult forest was substantial, with the success rate dropping from 15/15 to 5/15. Additionally, overall flight stability decreased markedly, and the number of failures not caused by leaves or direct collisions with obstacles increased significantly.

Flight mission completion times were significantly reduced with the optimized system compared to the original system. In the medium forest, the average completion time decreased from 89.2 s to 78.7 s at a target flight speed of 1 m/s, and dropped further to 41.3 s at 2 m/s. In the difficult forest, completion times were reduced from 144.3 s to 86.5 s at 1 m/s and to 56.8 s at 2 m/s. The primary reason for these reductions was the elimination of in-place zigzag maneuvers present in the original system. Correspondingly, $t_{\text{extra}}$ decreased from 12.5 s to 3.2 and 4.0 s in the medium forest for target flight velocities of 1 m/s and 2 m/s, respectively. In the difficult forest, the reductions were even more noticeable, from 60.4 s to 9.7 and 12.0 s for the same target flight velocities. Additionally, the optimized system achieved a modest increase in average flight speed from 0.76 m/s to 0.80 m/s in the medium forest and from 0.70 m/s to 0.74 m/s in the difficult forest due to smoother planned trajectories which could be flown with slightly higher speeds. 

With the optimized system, flight mission completion times increased only slightly when moving from the medium to the difficult forest, primarily due to the greater number of obstacles to dodge and the emergency stop periods triggered by prolonged A* path searches near the cluttered goal area in the difficult forest. While the total duration of emergency stops was small for most flights, some lasted longer than 20 s.

There was a significant speed increase when the target flight speed was increased from 1 m/s to 2 m/s with the optimized system. In the medium forest, the average flying speed increased from 0.80 m/s to 1.48 m/s, and the point-to-point average speed increased from 0.76 m/s to 1.35 m/s. In the difficult forest, the increase in speed was from 0.74 m/s to 1.32 m/s and from 0.66 m/s to 1.06 m/s, respectively. In any case, the actual flight speed did not double, although the target flight speed doubled. The reason for this was that the increase in flight speed led to more emergency stops and less efficient routes taken due to shorter reaction time to previously occluded obstacles and more leaves being pushed up from the ground. 

\section{DISCUSSION}
\label{sec:discussion}

\subsection{ANALYSIS OF SYSTEM PERFORMANCE AND WAYS TO IMPROVE THE SYSTEM}

For an autonomous system to be considered a viable replacement for the current state-of-the-art under-canopy remote sensing solutions, it must demonstrate near-perfect reliability, even in dense forests, while achieving point-to-point flight speeds comparable to those of a human operator performing mobile remote sensing. Since a human operator can adapt to a wide range of scenarios, the success rate of human based data collection can be considered to be 100 \%. A realistic walking speed of a human conducting remote sensing in forested terrain is approximately 1--1.5 m/s. Furthermore, the system should exhibit almost no collisions with obstacles in order to minimize the risk of hardware damage and prevent harm to forest vegetation.

The optimized system demonstrated promising performance, especially at low target flight velocities. Success rates of 12/15 and 15/15 at a target flight speed of 1 m/s indicate that the optimized system is suitable for reliable under-canopy flights when loose leaves are not present on the ground and therefore cannot be lifted by the quadrotor propellers. However, since flying leaves significantly reduces system reliability, developing methods to filter leaves and other lightweight foliage from the point cloud would substantially improve robustness and enable operations across a wider range of forest environments. 

At a target flight speed of 1 m/s, the optimized system collided with branches four times. Although such collisions with obstacles were infrequent, further improvements in flight stability are necessary to minimize the collision probability. Moreover, despite the target flight speed of 1 m/s, the actual point-to-point flight speed was considerably lower. To achieve a flight speed more comparable to the human walking speed, the efficiency of the A* path search should be improved to reduce the number of emergency stops, particularly near cluttered goal locations. 

Increasing the target flight speed to 2 m/s increased both the average flying speed and point-to-point speed of flights to levels comparable to a human operator in both forests (1.48 m/s and 1.35 m/s in the medium forest and 1.32 m/s and 1.06 m/s in the difficult forest). However, there was a reduction in system reliability, particularly in the forest with high structural complexity, which makes increasing the target flight speed from 1 to 2 m/s inadvisable. In the difficult forest environment, the success rate dropped below 50 \%, and the number of collisions during flights increased substantially. Although the success rate in the medium forest remained comparable to that achieved at a target flight speed of 1 m/s, operating at higher speeds is still not recommended, as the number of collisions increased even in a less challenging environment in the medium forest. Based on these evaluations, we identified that there were two primary factors limiting the achievable operational speed of the system. In addition to the increased difficulty of recovering from a cloud of leaves rising from the ground around the quadrotor, the structural complexity and partial observability of real forest environments, combined with unstable flight at higher speeds, limit the operational speed of the system. Although the lidar is able to accurately reconstruct the three-dimensional structure within its current field of view, the environment beyond a few meters remains unknown. In dense, cluttered forests, previously occluded obstacles may become visible only after the platform has already passed nearby trees. This, combined with less stable flight with higher speeds, can lead to sudden obstacle detections, which may require a set of infeasible control commands to avoid collisions. Moreover, the reduced flight stability observed at higher velocities may degrade the quality of the collected remote sensing data, an effect that is likely to be especially pronounced when using cameras in addition to lidar. 

Overall, although the optimized system demonstrates promise, its reliability requires further improvement to be considered as a viable alternative to human-operated mobile remote sensing. Specifically, the failure rates when leaves are present on the ground, as well as the likelihood of collisions with obstacles, should be decreased. Additionally, the flight stability of the system at higher velocities should be increased. 

Since forests are heterogeneous in both density and structural complexity, and the optimized system exhibited varying performance as the target flight speed increased in more difficult environments, the ability to autonomously adapt the target flight speed based on the characteristics of upcoming forest patches would be beneficial. Such adaptation could be driven by the density of the obstacle map and the amount of estimated free space around the planned path. When the obstacle density is low and plenty of free space is available, the target flight speed could be increased. Conversely, upon detecting denser forest regions, the target speed could be reduced to maintain flight safety and reliability. For example, Zhao et al. \cite{zhao10582409} proposed a reinforcement learning based algorithm for adaptive flight speed control designed to operate alongside traditional path planning and control methods. Although the approach was originally developed for camera-based systems, it relies on an occupancy grid as input, making it directly applicable to lidar-based systems as well. In the original study, the algorithm was evaluated in both indoor and forest environments. Based on the supplementary video, in an approximately medium-complexity forest, the system successfully adapted its flight speed between 0.2 and 3.6 m/s depending on the local environmental complexity. These results of Zhao et al. \cite{zhao10582409} suggest that incorporating adaptive flight speed control could further improve the reliability, flight stability, and efficiency of the optimized system. 

Another potential way to improve the reliability, stability, and decrease the probability of collisions is to introduce a safety margin around obstacles, within paths that are not planned unless alternative routes to the goal do not exist. Since A* by definition seeks an optimal path, it tends to generate trajectories that pass obstacles as close as permitted by the obstacle inflation radius in order to minimize path length. Such close proximity increases the risk of collision in the presence of trajectory tracking disturbances or a distance estimation error, whereas simply increasing the inflation radius completely prevents the system from flying through tight gaps. By making the inflation radius adaptive, e.g., with a soft safety margin, the system could favor safer trajectories whenever possible, while still allowing passages through tighter, higher-risk gaps when necessary. Flying further away from the detected obstacles decreases the probability of situations where suddenly appearing obstacles make the control commands to avoid the obstacles infeasible.

\subsection{COMPARISON TO OTHER SYSTEMS}

Comprehensive comparisons with most existing under-canopy flight systems are challenging due to variations in testing conditions and metrics, and their reporting. The recommended evaluation metrics are described in detail in \autoref{subsec:standardizedSetup}.  

Among the studies reviewed in \autoref{sec:relatedWork}, only the two works by Karjalainen et al. \cite{karjalainen2023drone, Karjalainen03122025} reported all key evaluation metrics, including forest density, success rate, flight distance, and target flight speed. In the study of Karjalainen et al. \cite{karjalainen2023drone}, nine out of 19 flights were successful in a forest with a density ranging from 1650 to 2380 trees/ha and a high number of low-hanging branches. The target flight speed was 1.0 m/s, and the flight distances ranged from 35 to 80 m. In the study of Karjalainen et al. \cite{Karjalainen03122025} the flights were conducted in forests using the same forest complexity criteria applied in this study \cite{liang2019forest}. The test environments were classified as medium complexity, with a density of 650 trees/ha, and difficult complexity, with a density of 2000 trees/ha. Both forests contained a high number of low-hanging branches and other understory foliage. In the medium forest, all flights were successful, whereas in the difficult forest, eight out of nine flights were successful. The target flight speed was 1.0 m/s, with flight distances ranging from 34 to 36 m in the medium forest and 42 m in the difficult forest. 

The optimized system showed clearly better performance than the original system used by Karjalainen et al. in \cite{karjalainen2023drone}, achieving a success rate of 100 \% in the difficult forest compared with a success rate of 47 \% reported in that study. In comparison with the system presented by Karjalainen et al. in \cite{Karjalainen03122025}, the performance difference is smaller. In the medium forest, the reliability of the optimized system was lower (80\% versus 100 \%), whereas in the difficult forest it was higher (100\% versus 89 \%). It should be noted, however, that the flights conducted by Karjalainen et al. were shorter (42 m) than those in this study (60 m), which likely made their missions easier to complete. Nevertheless, the higher reliability of the optimized system in the difficult forest, together with the observation that all failures in the medium forest were caused by the clouds of leaves, suggests that the optimized system is better suited for denser forests and, consequently, a wider range of forest types in terms of reliability. Leaves were thrust from the ground during the experiments conducted by Karjalainen et al. It was reported that the leaves did not affect the localization accuracy. Since no issues with leaves relating to the flight performance were reported, it is likely that the camera-based system was not affected by leaves being thrust from the ground. A direct comparison of efficiency is not possible, as no actual flight speeds, point-to-point flight speeds, or flight mission completion times were reported in the referenced studies. Overall, the optimized system presented in this work appears to perform more robustly across varying forest environments, at least in scenarios where large amounts of loose leaves are not present on the ground. 

Among the reviewed approaches that did not report the proposed minimum evaluation criteria, the system of Ren et al. \cite{ren2025safety} appears to be the most promising. The combination of very high actual flight speeds (5--12.5 m/s) and a 100 \% success rate across all eight flights demonstrates that the system is capable of reliably executing high speed flights. However, its performance in truly dense forest environments remains unclear, as the reported flights were conducted primarily in open fields and in forested areas characterized by large openings, wide gaps between trees, and trails that likely facilitated traversal through the forest section. 

Flying in dense forests imposes different requirements than flying at high speed in open areas. Open-field flight emphasizes long-range obstacle detection and planning smooth, feasible trajectories around sparse obstacles at high velocities. In contrast, dense forest flight demands rapid reactions to occluded obstacles, the ability to safely navigate tight gaps between trees and branches, robust responses to dynamic elements such as moving branches and foliage influenced by wind and quadrotor downwash. Owing to these differing operational demands, the performance of the system in the study of Ren et al. \cite{ren2025safety} under dense forest conditions cannot be reliably inferred from the reported experiments. The person tracking experiment presented by Ren et al. suggests that their system may be capable of operating in dense forest environments. Nevertheless, the limited number of trials, insufficient reporting of forest characteristics, and the apparent absence of low-hanging branches in the test environment hinder a rigorous assessment of system performance. Overall, while it is possible that the system of Ren et al. could outperform the optimized system presented in this study even in dense forests, the lack of comprehensive and comparable evaluation data makes accurate performance comparisons difficult.   

Based on the results of this study, the original system employed here, which closely follows the approach by Liu et al. \cite{liu2023integrated}, exhibited different performance characteristics in dense forest environments compared to the lower-complexity forest scenarios reported in their work. Liu et al. demonstrated the effectiveness of their approach by successfully completing an approximately 60 m autonomous flight with a target speed of 6 m/s. In the present study, when operating in medium-density forest conditions, the original system was not able to complete a 60 m  flight even at a lower target speed of 2 m/s, illustrating the substantially increased challenges posed by denser and more cluttered environments.   

The comprehensive experimental campaign conducted in this work revealed several behaviors that were not explicitly addressed in earlier evaluations. These included oscillatory flight behavior in dense environments, challenges related to height estimation when initiating flights from sloped terrain, and limitations in NaN handling during the planning process. The NaN-related issue was associated with implementation-level error handling rather than the underlying algorithm itself and originated from the publicly available code by Liu and Ren \cite{IPCsoftware}. Importantly, the core planning and control framework introduced by Liu et al. proved to be highly capable, and after the improvements introduced in this work, it enabled robust and reliable performance in highly cluttered forest environments.

Overall, these findings highlight the importance of extensive, standardized, and systematically varied testing in complex real-world environments, where subtle implementation details and environment-specific effects can only be revealed through repeated and consistent experimental evaluation.

\subsection{STANDARDIZED EXPERIMENTAL SETUP}
\label{subsec:standardizedSetup}

Based on the literature review in \autoref{sec:relatedWork}, the testing schemes and reporting of autonomous flight systems for forest flight systems require improvement within the scientific community. Currently, many experimental setups lack important details. First, forest test areas are often insufficiently described. Of the ten studies with real-world forest flights reviewed here, only four \cite{huan2024uncertainty,han2025dynamically,karjalainen2023drone,Karjalainen03122025} reported forest densities numerically. In the remaining studies, tree density was estimated visually in this work based on the images and videos provided, varying from low (less than 700 trees/ha) to high (over 2000 trees/ha). Similarly, the number of low-hanging branches and other understory foliage was reported in only two studies \cite{delcol2025autonomous,karjalainen2023drone}, while only one study \cite{Karjalainen03122025} applied similar, quantitative forest evaluation criteria \cite{liang2019forest} used in this work. In some cases, test flights were conducted along routes with few obstacles or existing paths, which likely simplified traversal. 
Second, many studies report only a single successful flight or omit the success rate entirely, making it impossible to assess reliability. Only four of the studies \cite{ren2025safety,delcol2025autonomous,karjalainen2023drone,Karjalainen03122025} reported the success rate. Third, the number of collisions that did not result in failures was rarely reported, appearing in only one study \cite{delcol2025autonomous}, although some studies may have omitted this detail because no collisions occurred. Finally, flight mission completion time or the point-to-point flight speed was reported in only one study \cite{campos2021autonomous}, whereas actual flight speed was reported in five studies\cite{liu2023integrated,ren2025safety,jacquet2025neural,campos2021autonomous,han2025dynamically}.

Since testing setups for forest flight experiments vary widely across studies, a more standardized testing framework is proposed. The experimental setup and its reporting are divided into four categories: (1) metrics related to the test forest environments and flight parameters, (2) metrics for system reliability, (3) metrics for system efficiency, and (4) metrics for location accuracy. In addition to being comprehensive, the proposed setup is designed to remain practical and not overly laborious.  An overview of the proposed setup is presented in \autoref{tab:experimentalSetup}.

(1) Metrics related to the test forests and flight parameters: The complexity of a forest is comprised of two major factors: tree density and the amount of branches and other foliage. Low-hanging branches and dense understorey are particularly critical for the difficulty of low-altitude flights, so these variables should always be reported. For example, the forest evaluation criteria proposed by Liang et al. \cite{liang2019forest} could be used to determine forest complexity. In addition to these structural characteristics, the presence of anomalies, such as leaves on the ground, should also be documented. 

In addition to a formal description of the test environments, visual representations of the test forests should be provided. Rather than relying solely on single images, researchers are encouraged to publish supplementary materials alongside their articles. Such materials could include flight-test videos, point clouds, or 3D models of the forests. 

Careful consideration should be given to the selection of test forests and flight paths. The flight path should not be chosen so that there are no obstacles on the direct path between the start and the goal. In addition, there should not be any parallel paths that the autonomous flying system could utilize to easily dodge the trees. Ideally, flights should be conducted in multiple forest environments with varying levels of complexity to provide a comprehensive assessment of the system performance. At least one test forest should be highly complex, with over 2000 trees/ha. Since a robust forest-flying system should operate reliably in all forest environments, testing in dense forests with low-hanging branches and foliage is essential. 


(2) Metrics for system reliability:
Reliability should be primarily assessed through success rates, which require multiple test flights rather than single demonstration. Flights conducted in different forest environments and at multiple target velocities are encouraged to evaluate reliability across varying conditions. In addition to success rates, qualitative observations, such as failure causes, contacts to obstacles not leading to a failure, and ineffective behaviors, such as emergency stops, should also be reported.     

(3) Metrics for system efficiency:
The results of the flight tests should be reported in sufficient detail. Relevant parameters include the total flight distances and the realized flight speed, rather than only the commanded target speed. To enable assessment of the route efficiency, either the mission completion time or point-to-point average flight speed should also be reported.


(4) Metrics for location accuracy:
Whenever relevant, the localization accuracy of the proposed system should be demonstrated. Ideally, this is done by collecting ground truth data along the autonomously flown path and comparing the estimated positions of the system to the ground truth, for example, using RMSE between the estimated and reference trajectories. Obtaining an accurate reference trajectory in forested environments can be challenging. Real-time kinematic (RTK) GNSS is attractive in open air but often infeasible under dense canopy. Various alternative methods have been used in the literature. For example, Karjalainen et al. \cite{karjalainen2024autonomous} estimated the reference trajectory photogrammetrically using a structure-from-motion technique with ground control points measured by a robotic tachymeter. For rapid testing, it may be sufficient to demonstrate localization accuracy through simpler means, such as manually flying the UAV and measuring end-point drift of the position estimate, as done by Karhunen et al. \cite{isprs-archives-XLVIII-2-W11-2025-153-2025}. The rapid testing method can be further enhanced by placing markers at the start and end locations of the test areas for local trajectory alignment across all flights. This would allow more robust trajectory analysis, as the flown trajectories could be registered and plotted within the point cloud of the test area. 

However, for comprehensive analysis, full trajectory estimation is recommended. Relying solely on widely used benchmark datasets, such as Mulran \cite{gskim-2020-mulran}, EuRoC MAV \cite{Burri25012016}, or KITTI \cite{Geiger2012CVPR}, is insufficient, since these either lack forest flight data or are not collected with UAVs. In some close-range under-canopy studies \cite{faitli2023realtime,wang2024benchmarking}, reference trajectories are determined offline by aligning mobile lidar point clouds to a high-quality terrestrial laser scanning (TLS) point cloud using an Iterative Closest Point (ICP) algorithm \cite{besl1992ICP}.  Since many robotic under-canopy UAV systems already record point clouds, this method is a practical option when other ground truth methods are infeasible and a survey-grade TLS reference is available.

\begin{table*}
    \caption{Overview of the proposed experimental setup.}
    \label{tab:experimentalSetup}
    \begin{tabularx}{\textwidth}{@{} p{4.5cm} *{1}{X} @{}}
    \toprule
    Forest metrics & Forest characterization by density and understory vegetation (according to \cite{liang2019forest}): easy (less than 700 trees/ha, minimal understory), medium (approx. 1000 trees/ha, sparse understory), difficult (approx. 2000 trees/ha, dense understory); amount of low-hanging branches; presence of anomalies such as leaves on the ground; visual representation of the test forest; at least one difficult test forest included \\
    Reliability metrics & Multiple flights, success rate, multiple environments, multiple flight speeds, and qualitative results such as reasons for failure, contacts to obstacles, observed ineffectivenesses \\
    Efficiency metrics & Flight distance, target flight speed, real flight speed, and point-to-point flight speed or flight mission completion time \\
    Localization accuracy metrics & Manual flight localization demonstration and markers at the starting point and the goal point or comparison to ground truth (photogrammetric structure-from-motion or point cloud matching to a high-quality point cloud) \\
    \bottomrule
    \end{tabularx}
\end{table*}

The proposed experimental setup is applicable to different frameworks used in under-canopy navigation, including vision- and lidar-based systems; indeed, some previous studies have already employed subsets of the proposed metrics. If the experimental setup presented above were applied to novel autonomous under-canopy algorithms and solutions, it would facilitate meaningful comparisons of different autonomous flying systems based on literature alone. Ideally, all systems would be tested at the same location and time to allow direct performance comparisons. Since this is rarely feasible, adopting the standardized setup proposed in this paper would substantially improve the comparability of different frameworks.

Simulation can serve a valuable preliminary testing tool. However, for under-canopy quadrotors, relying solely on simulations, at least with current state-of-the-art platforms, does not accurately reflect real-world performance. The complexity of forests, combined with the inherently chaotic nature of the natural environments, often makes simulations unrepresentative of actual autonomous flights. Therefore, a well-documented real-world experiments remain essential for demonstrating system performance and reliability.
 
Beyond improving comparability, extensive experimentation also enables system refinement. Based on tests conducted with the original system, performance was significantly enhanced through a series of modifications. The optimized system achieved a 100 \% success rate in the selected medium and dense forest environments, with a few exceptions caused by loose flying leaves. These experiments highlighted a new, remaining challenge for lidar-based autonomous flying quadrotors: handling loose foliage, which will require further development in the future.

\section{CONCLUSION}
\label{sec:conclusion}

In this paper, a standardized experimental setup is proposed for autonomous forest flight experiments, as current experimental practices vary widely in the literature. Due to insufficient rigor in both experimental design and reporting, accurate performance evaluation and comparison of existing systems remain challenging. To address this gap, the experimental setup proposed in this study is designed to be comprehensive and systemically varied, while remaining practical and not overly laborious to implement. 

The proposed standardized setup comprises four categories: forest metrics, system reliability, system efficiency, and localization accuracy. The setup was demonstrated through a series of experiments conducted on a quadrotor system implementing a state-of-the-art open-source lidar-based autonomous flight framework. The framework was deployed on a miniaturized forest UAV equipped with a lightweight lidar sensor, resulting in a total system weight of 1.2 kg (excluding batteries). Based on a systematic analysis of the experimental results, the original system was enhanced. The performance of the optimized system was then validated through comparisons with the original implementation and with relevant autonomous under-canopy flight systems reported in the literature.

The optimized system showed significantly improved performance compared to the original implementation. The reliability of the system, measured in terms of flight success rate, number of collisions with surrounding obstacles, and other undesirable behaviors, improved substantially. The flight efficiency of the system was also improved as indicated by reduced flight mission completion times and increased point-to-point and true flight speeds. Although direct comparisons with other systems remain challenging due to incomplete, inconsistent, or differing evaluation metrics reported in the literature, the results obtained are comparable to or outperform those reported in studies where meaningful comparisons are feasible. 

Several avenues for further improvement were identified. Most importantly, these include improved handling of lightweight foliage, such as leaves, adaptive velocity control to balance efficiency and reliability across varying forest densities, and enhanced safety margins around obstacles to favor safer trajectories when possible. Overall, the proposed experimental setup offers a practical means to improve literature-based comparability and to support systematic performance improvement of autonomous under-canopy UAV systems.


In the future, the aim is to deploy the autonomously navigating under-canopy UAV as a less labor-intensive and scalable alternative for collecting remote sensing data in forests. Equipped with additional sensors, such as cameras, the system has a range of potential applications, such as stem curvature and quality measurements of trees, forest health estimation, and biodiversity monitoring.

\bibliographystyle{IEEEtran}
\bibliography{article}

\section*{APPENDIX}
\label{sec:appendix}
This Appendix section presents a detailed overview of the individual test flights conducted in this study. \cref{tab:paloheinaOrgMed,tab:paloheinaOrgMed2ms} present the individual flights conducted with the original system in the medium difficulty forest with a target flight speed of 1 m/s and 2 m/s, respectively. \cref{tab:paloheinaOrgDif} presents the individual flights conducted with the original system in the difficult forest with a target flight speed of 1 m/s. \cref{tab:paloheinaOptMed,tab:paloheinaOptMed2ms} present the individual flights conducted with the optimized system in the medium forest, with a target flight speed of 1 m/s and 2 m/s, respectively. \cref{tab:paloheinaOptDif,tab:paloheinaOptDif2ms} present the individual flight conducted with the optimized system in the difficult forest with a target flight speed of 1 m/s and 2 m/s, respectively. The reasons for failures are presented for the failed flights in parentheses under the "Success" column with the following failure meanings: "Tree": collision with a tree, "Leaves": a cloud of leaves thrust from the ground, "NaN": NaNs generated during SFC generation, and "Unstable": unstable flying causing a failure. The point-to-point average speed, average flight speed, and $t_{\text{extra}}$ were measured only for successful flights. An "x" indicates that no data were collected.

\begin{table*}[tb]
	\centering
	\caption{Overview of the flights flown with the original system with a target flight speed of 1 m/s in the medium forest.}
	\label{tab:paloheinaOrgMed}
	\begin{tabularx}{\textwidth}{@{} l *{8}{C} c @{}}
	  \toprule
	  Flight & Success & Flight end point & Collisions & Aggressive leaf dodges & Flight time & Point-to-point average speed & Average flight speed & $t_{\text{extra}}$\\
    \midrule
	1 & Yes & (55.96, -0.45, 1.45) & 0 & 1 & 93.5 s & 0.62 m/s & 0.81 m/s & 24.4 s\\ 
    \rowcolor{lavender}
    2 & Yes & (58.14, -2.96, 0.83) & 0 & 2 & 102.0 s & 0.57 & 0.68 & 16.4 s\\
    3 & No (Leaves) & (8.76, -0.98, 1.16) & 1 & 5 & x & x & x & x\\ 
    \rowcolor{lavender}
    4 & Yes & (59.45, -2.04, 0.96) & 1 & 2 & 82.0 s & 0.73 & 0.75 & 2.7 s\\ 
    5 & Yes & (59.54, -1.93, 0.94) & 1 & 1 & 80.1 s& 0.74 & 0.80 & 5.6 s\\ 
    \rowcolor{lavender}
    6 & No (Tree) & (21.76, 0.08, 1.40) & 1 & 0 & x & x & x & x \\ 
    7 & No (NaN) & (42.13, -1.96, 1.39) & 0 & 0 & x & x & x & x \\ 
    \rowcolor{lavender}
    8 & Yes & (59.54, -2.00, 1.00) & 0 & 2 & 79.3 s& 0.75 & 0.83 & 7.5 s\\ 
    9 & Yes & (59.97, -2.01, 0.98) & 0 & 0 & 80.3 s& 0.75 & 0.79 & 4.3 s\\ 
    \rowcolor{lavender}
    10 & No (Leaves) & (29.81, -2.13, 1.42) & 0 & 4 & x & x & x & x \\ 
    11 & Yes & Unknown & 1 & 1 & $\approx$79 s & $\approx$0.76 & unknown & unknown\\
    \rowcolor{lavender}
    12 & No (Unstable) & (20.17, 0.32, 1.73) & 0 & 0 & x & x & x & x \\
    13 & Yes & (57.58, -3.02, 0.95) & 0 & 0 & 75.2 s & 0.77 & 0.79 & 2.2 s\\
    \rowcolor{lavender}
    14 & Yes & (57.88, -2.89, 0.92) & 6 & 1 & 133.8 s & 0.43 & 0.60 & 37.2 s\\
    15 & Yes & (58.04, -2.79, 0.92) & 1 & 1 & 87.0 s & 0.67 & 0.78 & 12.5 s\\
    \bottomrule
	\end{tabularx}
\end{table*}

\begin{table*}[tb]
	\centering
	\caption{Overview of the flights flown with the original system with a target flight speed of 2 m/s in the medium forest.}
	\label{tab:paloheinaOrgMed2ms}
	\begin{tabularx}{\textwidth}{@{} l *{8}{C} c @{}}
	  \toprule
	  Flight & Success & Flight end point & Collisions & Aggressive leaf dodges  & Flight time & Point-to-point average speed & Average flight speed & $t_{\text{extra}}$\\
    \midrule
    1 & No (Unstable) & (17.22, -0.60, 0.44) & 1 & 3 & x	& x	& x	& x \\
    \rowcolor{lavender}
    2 & No (Unstable) & (10.74, 2.75, 0.08) & 0 & 0	& x	& x	& x & x \\
    3 & No (Unstable) & (12.90, 0.41, 0.42) & 0 & 1	& x	& x & x	& x \\
    \bottomrule
    \end{tabularx}
\end{table*}

\begin{table*}[tb]
	\centering
	\caption{Overview of the flights flown with the original system with a target flight speed of 1 m/s in the difficult forest.}
	\label{tab:paloheinaOrgDif}
	\begin{tabularx}{\textwidth}{@{} l *{8}{C} c @{}}
	  \toprule
	  Flight & Success & Flight end point & Collisions & Aggressive leaf dodges  & Flight time & Point-to-point average speed & Average flight speed & $t_{\text{extra}}$\\
    \midrule
	1 & Yes & (55.96, -0.45, 1.45) & 0 & 0 & 94.8 s& 0.59 & 0.64 & 7.4 s\\
    \rowcolor{lavender}
    2 & Yes & (58.50, 1.38, 0.97) & 0 & 0 & 101.3 s& 0.58 & 0.75 & 23.3 s\\ 
    3 & No (NaN) & (45.17, -2.58, 0.00) & 1 & 0 & x & x & x & x\\ 
    \rowcolor{lavender}
    4 & Yes & (60.00, -2.74, 0.93) & 0 & 0 & 194.7 s& 0.31 & 0.63 & 99.3 s\\ 
    5 & No (Tree) & (43.27, -0.35, 0.94) & 1 & 0 & x & x & x & x\\ 
    \rowcolor{lavender}
    6 & Yes & (57.26, -0.95, -0.06) & 7 & 0 & 175.4 s& 0.33 & 0.70 & 93.6 s\\
    7 & No (Unstable) & $\approx$(30, 2, 1) & 1 & 0 & x & x & x & x\\ 
    \rowcolor{lavender}
    8 & Yes & (55.87, -3.70, 1.38) & 1 & 1 & 157.6 s & 0.35 & 0.72 & 79.8 s\\
    9 & No (NaN) & (41.80, 0.21, 1.40) & 0 & 0 & x & x & x & x\\ 
    \rowcolor{lavender}
    10 & No (Unstable) & (44.24, -2.04, 0.01) & 0 & 0 & x & x & x & x\\ 
    11 & No (Tree) & (34.48, -1.63, 0.35) & 1 & 0 & x & x & x & x\\ 
    \rowcolor{lavender}
    12 & No (NaN) & (29.81, 0.09, 0.16) & 0 & 0 & x & x & x & x\\ 
    13 & No (Tree) & (34.17, -2.06, 0.24) & 1 & 0 & x & x & x & x\\
    \rowcolor{lavender}
    14 & Yes & (63.26, 0.75, -0.16) & 9 & 0 & 142.0 s & 0.45 & 0.76 & 58.7 s\\
    15 & No (NaN) & (48.15, 1.13, 0.37) & 0 & 0 & x & x & x & x\\ 
    \bottomrule
    \end{tabularx}
\end{table*}

\begin{table*}[tb]
	\centering
	\caption{Overview of the flights flown with the optimized system with a target flight speed of 1 m/s in the medium forest.}
	\label{tab:paloheinaOptMed}
	\begin{tabularx}{\textwidth}{@{} l *{8}{C} c @{}}
	  \toprule
	Flight & Success & Flight end point & Collisions & Aggressive leaf dodges  & Flight time & Point-to-point average speed & Average flight speed & $t_{\text{extra}}$\\
    \midrule
	1 & Yes & (60.07, -0.14, 1.04)	& 0 & 1 & 75.6	& 0.79	& 0.82 & 2.0 s\\ 
    \rowcolor{lavender}
    2 & Yes & (59.98, 0.01, 0.98) & 0 & 3 & 80.2	& 0.75	& 0.78 & 3.7 s\\ 
    3 & No (Leaves) & (11.15, 0.93, 0.00) & 0 & 1 & x & x & x & x\\ 
    \rowcolor{lavender}
    4 & No (Leaves) & (11.31, 0.85, 0.10) & 0 & 1 & x & x & x & x \\ 
    5 & Yes & (60.00, -0.01, 1.46) & 0 & 1 & 77.4	&  0.78	& 0.83 & 4.8 s\\
    \rowcolor{lavender}
    6 & Yes & (59.99, -0.02, 1.44) & 0 & 0	&  75.6	& 0.79	& 0.84 & 4.0 s\\
    7 & Yes & (60.01, -0.01, 1.45) & 0 & 0 & 76.1	& 0.79	& 0.80 & 0.7 s\\
    \rowcolor{lavender}
    8 & Yes & (59.86, 0.01, 1.46) & 0 & 0 & 76.8	& 0.78	& 0.81 & 2.8 s\\
    9 & Yes & (59.99, -0.02, 1.42) & 0 & 0 & 77.6	& 0.77	& 0.79 & 1.2 s\\
    \rowcolor{lavender}
    10 & Yes & (60.01, -0.03, 1.47) & 0 & 1	& 77.1	& 0.78	& 0.80 & 1.9 s\\ 
    11 & Yes & (60.13, 0.01, 1.42) & 1 & 1 &  93.1	& 0.65	& 0.70 & 6.7 s\\ 
    \rowcolor{lavender}
    12 & No (Leaves) & (35.23, -1.29, 0.77) & 0 & 2 & x & x & x & x\\ 
    13 & Yes & (60.14, -0.00, 1.48) & 0 & 1	& 76.9	& 0.78	& 0.81 & 3.1 s\\
    \rowcolor{lavender}
    14 & Yes & (60.10, 0.00, 1.47) & 0 & 0 &  76.8	& 0.78	& 0.81 & 2.5 s\\
    15 & Yes & (60.13, -0.03, 1.46) & 1& 2	& 81.7	& 0.74 &	0.78 & 5.0 s\\ 
    \bottomrule
    \end{tabularx}
\end{table*}

\begin{table*}[tb]
	\centering
	\caption{Overview of the flights flown with the optimized system with a target flight speed of 2 m/s in the medium forest.}
	\label{tab:paloheinaOptMed2ms}
	\begin{tabularx}{\textwidth}{@{} l *{8}{C} c @{}}
	  \toprule
	  Flight & Success & Flight end point & Collisions & Aggressive leaf dodges  & Flight time & Point-to-point average speed & Average flight speed & $t_{\text{extra}}$\\
    \midrule
    1 & Yes & (59.95, -0.01, 1.47) & 0 & 0 & 39.6	& 1.51	& 1.58 & 1.8 s\\
    \rowcolor{lavender}
    2 & Yes & (59.97, -0.02, 1.47) & 0 & 0	& 40.4	& 1.48 & 1.53 & 1.2 s\\
    3 & Yes & (60.04, -0.01, 1.47) & 0 & 0 & 44.7 & 1.34	& 1.44 & 3.2 s\\
    \rowcolor{lavender}
    4 & Yes & (59.94, -0.03, 1.46) & 0 & 0	& 40.9	& 1.46	& 1.53 & 1.8 s\\
    5 & Yes & (59.93, 0.01, 1.52) & 2 & 1	& 49.4 & 1.21	& 1.45 & 8.0 s\\
    \rowcolor{lavender}
    6 & Yes & (59.96, -0.01, 1.53) & 0 & 1	& 41.7	& 1.44 & 1.53 & 2.5 s\\
    7 & No (Leaves) & (34.15, 0.57, 0.91) & 0 & 1	& x & x	& x & x \\
    \rowcolor{lavender}
    8 & Yes & (59.98, -0.02, 1.48) & 0 & 0	& 43.5	& 1.38 & 1.54 & 4.5 s\\
    9 & No (Leaves) & (16.45, 4.10, 0.34) & 1 & 1 & x	& x	& x	& x \\
    \rowcolor{lavender}
    10 & No (Unstable) & (59.94, -0.42, 0.43) & 1 & 4 & x	& x	& x	& x \\
    11 & Yes & (60.02, 0.04, 1.52) & 0 & 1	& 42.3	& 1.42	& 1.54 & 3.3 s\\
    \rowcolor{lavender}
    12 & Yes & (59.97, 0.03, 1.58) & 1 & 3	& 47.4	& 1.26	& 1.46 & 6.3 s\\
    13 & Yes & (59.93, 0.03, 1.58) & 4 & 3	& 51.7	& 1.16	& 1.42 & 9.5 s\\
    \rowcolor{lavender}
    14 & Yes & (59.97, 0.06, 1.51) & 0 & 1	& 42.1	& 1.42	& 1.53 & 2.8 s\\
    15 & Yes & (60.08, 0.04, 1.51) & 0 & 0	& 52.8	& 1.14	& 1.23 & 3.9 s\\
    \bottomrule
    \end{tabularx}
\end{table*}

\begin{table*}[tb]
	\centering
	\caption{Overview of the flights flown with the optimized system with a target flight speed of 1 m/s in the difficult forest.}
	\label{tab:paloheinaOptDif}
	\begin{tabularx}{\textwidth}{@{} l *{8}{C} c @{}}
	  \toprule
	  Flight & Success & Flight end point & Collisions & Aggressive leaf dodges  & Flight time & Point-to-point average speed & Average flight speed & $t_{\text{extra}}$\\
    \midrule
    1 & Yes & $\approx$ (55.2, 0.14, 0.93) & 0 & 0	& 76.6	& $\approx$ 0.72	& $\approx$ 0.78 &  $\approx$ 6.0 s\\
    \rowcolor{lavender}
    2 & Yes & (55.91, 0.33, 0.15) & 0 & 1	& 86.2	& 0.65 & 0.74 & 10.8 s\\
    3 & Yes & (56.37, -0.26, 0.37) & 0 & 0	& 79.2 & 0.71	& 0.77 & 5.6 s\\
    \rowcolor{lavender}
    4 & Yes & (55.86, -0.10, 0.45) & 0 & 3	& 81.6	& 0.68	& 0.78 & 9.9 s\\
    5 & Yes & (55.64, 0.20, 0.57) & 0 & 0	& 78.3 & 0.71	& 0.76 & 5.5 s\\
    \rowcolor{lavender}
    6 & Yes & (58.59, 1.40, 0.91) & 0 & 2	& 115.9	& 0.51 &	0.62 & 22.0 s\\
    7 & Yes & (55.95, 1.61 0.04) & 1 & 0	& 82.7	& 0.68	& 0.76 & 9.2 s\\
    \rowcolor{lavender}
    8 & Yes & (58.61, 1.45, 0.93) & 0 & 1	& 91.8	& 0.64	& 0.76 & 14.3 s\\
    9 & Yes & (55.49, -0.55, 1.19) & 0 & 1	& 80.9	& 0.69	& 0.77 & 3.7 s\\
    \rowcolor{lavender}
    10 & Yes & (56.63, -1.42, 0.87) & 0 & 0	& 82.9	& 0.68	& 0.77 & 9.0 s\\
    11 & Yes & (54.74, -0.40, 0.14) & 0 & 0	& 79.6	& 0.69	& 0.77 & 5.1 s\\
    \rowcolor{lavender}
    12 & Yes & (55.66, -0.18, 0.02) & 0 & 0 & 77.7	& 0.72	& 0.78 & 6.2 s \\
    13 & Yes & (58.60, 1.33, 0.96) & 0 & 0	& 80.0	& 0.73	& 0.79 & 6.0 s\\
    \rowcolor{lavender}
    14 & Yes & (58.54, 1.40, 0.93) & 0 & 1	& 107.3	& 0.55	& 0.65 & 16.7 s\\
    15 & Yes & (58.95, 1.05, 0.93) & 1 & 0	& 96.6	& 0.61	& 0.72 & 15.0 s\\
    \bottomrule
    \end{tabularx}
\end{table*}

\begin{table*}[tb]
	\centering
	\caption{Overview of the flights flown with the optimized system with a target flight speed of 2 m/s in the difficult forest.}
	\label{tab:paloheinaOptDif2ms}
	\begin{tabularx}{\textwidth}{@{} l *{8}{C} c @{}}
	  \toprule
	  Flight & Success & Flight end point & Collisions & Aggressive leaf dodges  & Flight time & Point-to-point average speed & Average flight speed & $t_{\text{extra}}$\\
    \midrule
    1 & No (Leaves) & (11.52, 1.05, 0.01) & 0 & 1 & x	& x	& x	& x \\
    \rowcolor{lavender}
    2 & Yes & (60.63, -0.03, 0.94) & 2 & 2	& 64.3 & 0.94 & 1.17 & 12.7 s\\
    3 & No (Unstable) & (22.71, -1.50, 0.12) & 0 & 0 & x	& x & x	& x \\
    \rowcolor{lavender}
    4 & No (Leaves) & (29.91, -0.23, 0.00) & 2 & 1	& x	& x & x	& x \\
    5 & No (Leaves) & (53.07, -0.08, 0.01) & 2 & 5	& x	& x & x	& x \\
    \rowcolor{lavender}
    6 & No (Tree) & (14.89, -1.02, 0.63) & 1 & 0	& x	& x & x	& x \\
    7 & No (Unstable) & (28.47, 1.55, 0.09) & 1 & 0 & x	& x & x	& x \\
    \rowcolor{lavender}
    8 & No (Tree) & (27.25, 0.92, 0.01) & 2 & 1	& x	& x & x	& x \\
    9 & Yes & (55.15, -0.46, 0.63) & 1 & 1 & 49.3	& 1.12	& 1.38 & 9.4 s\\
    \rowcolor{lavender}
    10 & No (Unstable) & (28.27, 0.36, 0.28) & 0 & 0 & x	& x & x	& x \\
    11 & Yes & (56.51, -1.06, 1.04) & 0 & 0 & 52.3	& 1.08	& 1.45 & 13.1 s\\
    \rowcolor{lavender}
    12 & No (Leaves) & (30.43, -0.32, 0.22) & 2 & 1	& x	& x & x	& x \\
    13 & No (Leaves) & (47.83, -2.19, 0.10) & 0 & 1	& x	& x & x	& x \\
    \rowcolor{lavender}
    14 & Yes & (56.60, 1.20, 0.99) & 0 & 2	& 42.0	& 1.35	& 1.56 & 5.7 s\\
    15 & Yes & (59.99, -0.46, 1.01) & 0 & 4	& 76.1	& 0.79	& 1.06 & 19.3 s\\
    \bottomrule
    \end{tabularx}
\end{table*}

\end{document}